\PassOptionsToPackage{table}{xcolor}
\documentclass[10pt,twocolumn,letterpaper]{article}

\usepackage{cvpr}              

\usepackage[accsupp]{axessibility}

\newcommand{\inlinesubsection}[1]{\noindent\textbf{#1}\hspace{1.8ex}}
\DeclareMathOperator*{\argmin}{arg\,min}

\newcommand{\method}{PEDRA}
\newcommand{\methoddef}{\method\ (PEdestrian Dynamics Realism Assessment)}
\newcommand{\emdsymb}{\mathbf{E}}

\definecolor{cvprblue}{rgb}{0.21,0.49,0.74}
\usepackage[pagebackref,breaklinks,colorlinks,allcolors=cvprblue]{hyperref}
\usepackage{multirow}
\usepackage{mathtools}
\usepackage{bm}

\def\paperID{17475} 
\def\confName{CVPR}
\def\confYear{2026}

\title{\method: Evaluating the Realism of Pedestrian Dynamics in Video Generation}

\author{Aaron Appelle and Jerome P. Lynch\\
Duke University\\
{\tt\small \{aaron.appelle, jerome.lynch\}@duke.edu}}

\begin{document}
\maketitle

\begin{abstract}
Pedestrian simulation traditionally relies on expert-tuned, hand-crafted models that limit scalability and generalization. Meanwhile, large-scale video generation models have achieved high visual realism across diverse settings, motivating exploration of their potential as general-purpose world simulators. Existing benchmarks primarily assess single-subject realism rather than scenes with multiple interacting people, leaving the plausibility of multi-agent dynamics in generated videos untested. We propose a rigorous evaluation protocol to benchmark text-to-video (T2V) and image-to-video (I2V) models as implicit simulators of pedestrian dynamics. For I2V, we leverage start frames from established datasets to enable direct comparison with ground truth videos, while for T2V we design a prompt suite covering varied crowd densities and interaction types. A key component is a method to reconstruct 2D bird’s-eye view trajectories from pixel-space without known camera parameters. Our analysis shows that leading models exhibit effective priors for plausible multi-agent behavior, though issues such as merging and disappearing pedestrians reveal limits to their physical consistency.
\end{abstract}

\section{Introduction}
\begin{figure}[t]
    \centering
    \includegraphics[width=\linewidth]{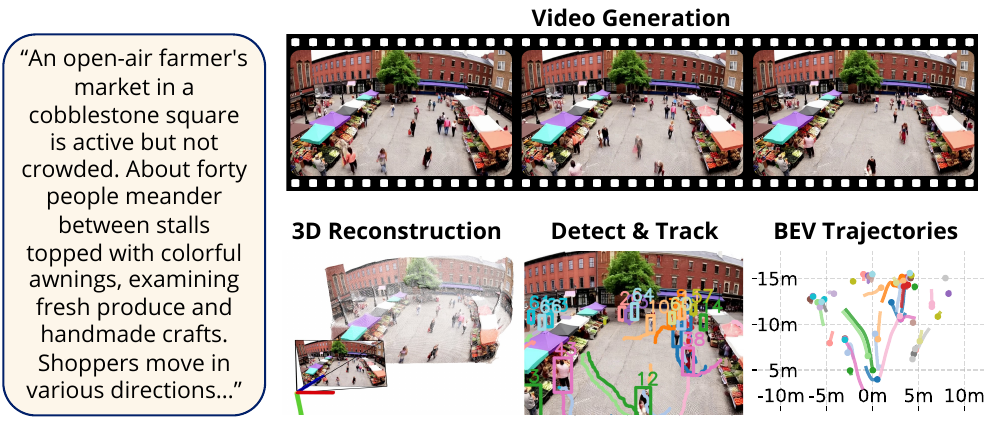}
    \caption{The \method\ method for text-to-video (T2V) models. From a text prompt, a video model generates a scene featuring pedestrian dynamics. We extract metric-scale trajectories from the synthetic video using 3D reconstruction to recover scene geometry and camera parameters, multi-object tracking to identify pedestrian paths in pixel-space, and projection of these paths into a unified bird's-eye view (BEV) coordinate system. The resulting trajectories are then analyzed for dynamic realism.}
    \label{fig:cover_pic}
\end{figure}

Realistic pedestrian simulation is essential for applications including autonomous driving \cite{fang_behavioral_2024,jiang_motiondiffuser_2023}, emergency planning \cite{barr_beyond_2024, xie_group_2021, sun_turkan_2020}, urban design \cite{feng_2016, mathew_2019, al-kodmany_crowd_2013}, human-robot interaction \cite{hossain_adr_2023,nocentini_survey_2019}, and computer graphics \cite{gomez_2024, nghiem_2021,armanto_2025,torrens_2024,wolinski_2014}. Modern crowd simulation frameworks integrate multiple components for global path planning, local trajectory modeling, and agent behavior \cite{martinez_2017,caramuta_2017}. However, practical adoption is hindered by significant limitations: defining and tuning simulations is a technically demanding manual process requiring domain expertise \cite{zhao_geraerts_2022}, and models often rely on heuristics or limited training data, leading to poor generalization \cite{kremyzas_towards_2016, papathanasopoulou_2024}.

Video generation models have made rapid advances in visual realism \cite{yang_cogvideox_2025,wan_wan_2025,kong_hunyuanvideo_2025,ju2025fulldit,chen_videocrafter2_2024,zheng2024open}, prompting research into their potential as general-purpose world simulators \cite{brooks_video_2024}. Initial studies on physics-based tasks such as rigid-body dynamics have shown promise \cite{leonardis_physgen_2025,montanaro_motioncraft_2024,zhang_physics-based_2024,2025veozeroshot}. Multi-agent pedestrian simulation presents a more complex testbed, as pedestrian behavior exhibits both physics-like properties \cite{castellano_2009,helbing1995social} and emergent social phenomena driven by human decision-making \cite{helbing2012social}. Given extensive internet-scale training data, these models may have learned latent spatiotemporal representations of multi-agent interactions, offering a new paradigm that could overcome generalization challenges \cite{appelle2025_icmlwm}.
However, existing video quality benchmarks \cite{huang_vbench_2024,zheng_vbench-20_2025} are not designed for scenes with many distant agents. The physical and behavioral plausibility of multi-agent interactions has not yet been systematically evaluated.

In this study, we introduce \methoddef, an evaluation protocol to assess the realism of crowd and pedestrian dynamics in videos generated by I2V and T2V models. For I2V, we benchmark generations conditioned on start frames from the ETH \cite{Lerner_ETH} and UCY \cite{Pellegrini_UCY} datasets, enabling direct comparison with ground truth videos. For T2V, we develop a prompt suite spanning diverse public scenes and social behaviors, structured along two axes: \emph{crowd density} (sparse, moderate, or crowded) and \emph{interaction type} (directional, multidirectional, or converging). We provide instructions for generating prompts using a large language model (LLM) and sample 5 videos per model for each prompt, totaling 900 videos per model.

Conducting the \method\ benchmark requires extracting 2D birds-eye-view pedestrian trajectories from the pixel-space of generated videos using a pre-trained multi-object tracker (MOT). For I2V, coordinate transformations are straightforward as known homographies are provided with the pedestrian trajectory datasets. For T2V with completely synthetic scenes, we introduce a method based on structure-from-motion (SfM) and metric depth-estimation to reconstruct trajectories without known camera parameters (Fig. \ref{fig:cover_pic}). We utilize Visual Geometry Grounded Transformer (VGGT) \cite{VGGT_Wang_2025_CVPR} to estimate the camera intrinsics and extrinsics, Depth Pro \cite{depthpro_bochkovskiy_2025} to estimate the metric-scale depth map of the generated scene, and then scale and align the pixel coordinates of the MOT bounding boxes to reconstruct the 2D trajectories.

We evaluate models using twelve metrics across three categories: \emph{trajectory kinematics, social interaction, and video fidelity}. Our analysis reveals that leading models possess an effective prior for plausible multi-agent behavior, successfully translating semantic prompts into varied crowd densities and interaction patterns, even replicating fundamental social phenomena. However, consistent failure modes also emerge. Pedestrians frequently disappear or merge together, and models often fail to render distinct individuals in large crowds. No single model excels across all scenarios, revealing trade-offs between scene fidelity, track consistency, and prompt adherence. This provides a performance baseline highlighting key areas for improvement in world modeling.
Our code is publicly available at \url{https://github.com/aaronappelle/PEDRA}.

\section{Related Work}
This work leverages video diffusion models (VDMs), which generate videos using the Diffusion Transformer (DiT) backbone \cite{peebles2023scalable} by performing denoising in a comrpessed latent space. A relevant history of VDMs is provided in the Appendix, Section \ref{sec:appdx_related}.\\

\inlinesubsection{Pedestrian Trajectory Prediction.}
Early work in this area relied on physics-inspired models \cite{helbing1995social,van2008reciprocal,bradley1993proposed}, which have been succeeded by deep learning methods that explicitly model complex social dynamics using LSTMs \cite{alahi_social_2016}, GANs \cite{gupta_social_2018}, and GNNs \cite{mohamed_social-stgcnn_2020, salzmann_trajectron_2020,shi_sgcn_2021}. State-of-the-art (SOTA) generative models based on transformers and diffusion now excel at synthesizing plausible, multi-modal trajectories \cite{yuan_agentformer_2021, gu_stochastic_2022, jiang_motiondiffuser_2023, chib_ms-tip_2024, ribeiro-gomes_motiongpt_2024}. Despite these advances, the predominant paradigm focuses on conditional short-term prediction of individual agents from past observations \cite{trajclip_2024, bae_continuous_2025}. These methods struggle to generalize to unseen environments or recreate multi-agent dynamics. This focus on single-agent short-term prediction distinguishes such approaches from holistic crowd simulation, which requires joint scene population and long-range multi-agent navigation \cite{bae_continuous_2025,seff_motionlm_2023,martinez_2017,caramuta_2017}.\\

\begin{figure*}[ht]
    \centering
    \includegraphics[width=\linewidth]{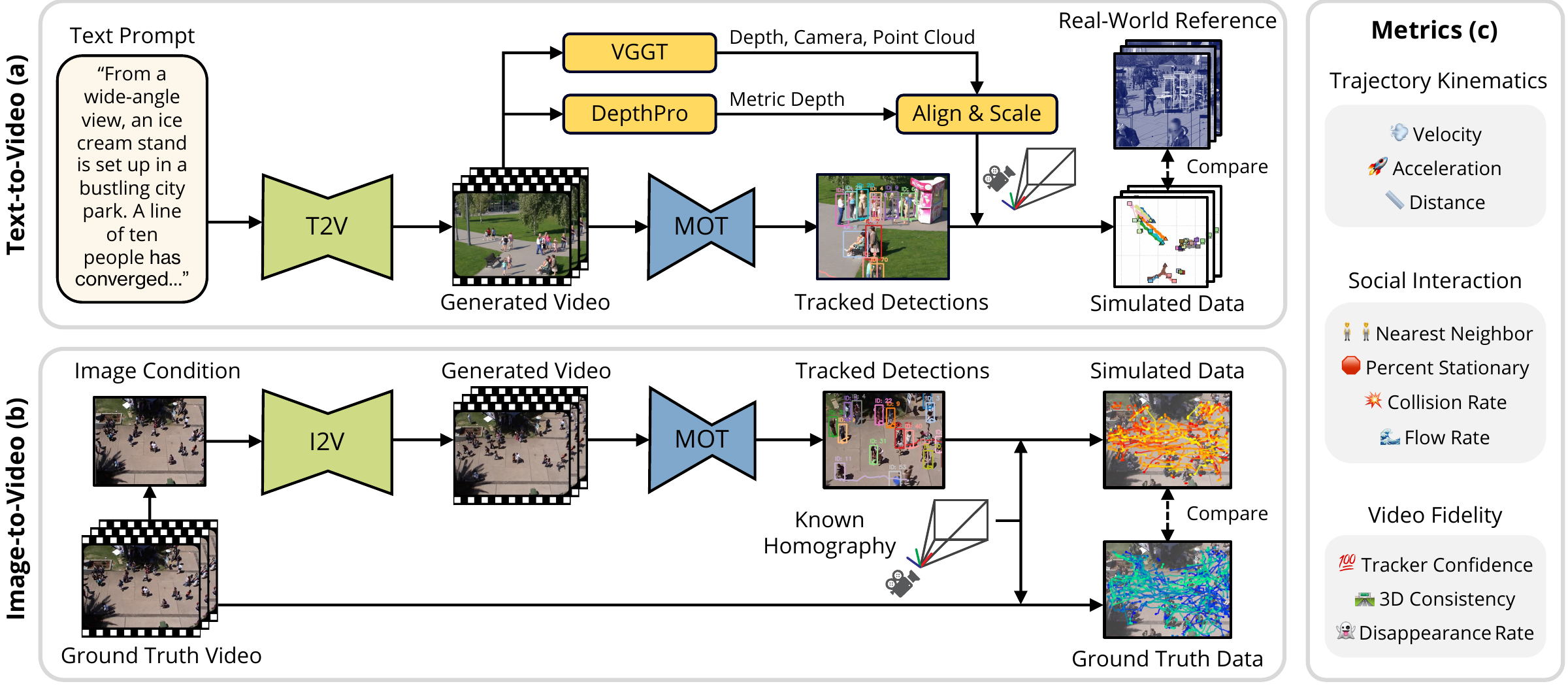} 
    \caption{\method\ consists of two primary benchmarks: (a) text-conditioned video generation leveraging 3D reconstruction to extract pedestrian trajectories, and (b) image-conditioned video generation, which can be directly compared against ground truth video using known image homography. Both benchmarks utilize a suite of metrics (c) to assess the realism of the resulting pedestrian simulation.}
    \label{fig:method_details}
\end{figure*}

\inlinesubsection{World Models.}
Recent work leveraging VDMs as world simulators \cite{brooks_video_2024} has focused on ensuring geometric consistency via explicit camera control \cite{ren2025gen3c,he_cameractrl_2025} and simulating physical dynamics for interactive scenarios \cite{zhang2024physdreamer,leonardis_physgen_2025,Li_2024_CVPR}. This concept is related to \emph{world models} in reinforcement learning, which are action-conditioned generative models of an agent's environment \cite{ha2018recurrent, Hafner2020Dream, hafner2021mastering}. Recent world models have incorporated generative architectures like diffusion models as their predictive core \cite{alonso2024diffusion, yang2024learning}, enabling applications in latent action learning \cite{bruce2024genie}, robotic grounding \cite{luo2025grounding}, and zero-shot policy transfer \cite{assran2025vjepa2}. These approaches typically adopt an ego-centric, single-agent perspective, operating within a different paradigm than the multi-agent crowd simulation setting considered in this work.\\

\inlinesubsection{Video Evaluation Benchmarks.}
The evaluation of VDMs has evolved from single-score metrics towards comprehensive benchmarks that decompose quality into hierarchical dimensions such as temporal consistency, action realism, and aesthetics \cite{huang_vbench_2024, liu2024evalcrafter,zheng2025vbench20}. Subsequent efforts go beyond visual fidelity to include physical plausibility, motion dynamics, commonsense reasoning, and compositionality \cite{zheng2025vbench20, Sun_2025_CVPR, liao2024DEVIL}. For scalability, recent approaches leverage dedicated evaluation models trained on large-scale human preference data \cite{he2024videoscore, mou2025gradeo}. While these frameworks address basic human-object or static interactions, they are not designed for scenes with multiple distant agents and do not systematically evaluate multi-person interaction dynamics. Most approaches rely heavily on vision-language models (VLMs) for evaluation rather than grounding assessments in strong real-world priors from pedestrian datasets, leaving a gap in understanding whether generated crowd behaviors exhibit physically and socially plausible properties.
\newcommand{\denS}{Sp.}
\newcommand{\denM}{Mo.}
\newcommand{\denC}{Cr.}
\newcommand{\intD}{Di.}
\newcommand{\intM}{Mu.}
\newcommand{\intC}{Co.}

\newcommand{\metricVelocity}{Velocity}
\newcommand{\metricAcceleration}{Acceleration}
\newcommand{\metricDistance}{Distance}
\newcommand{\metricDTW}{Path Error}
\newcommand{\metricDiversity}{Path Diversity}
\newcommand{\metricInternalDiversity}{Internal Diversity}
\newcommand{\metricCollision}{Collision}
\newcommand{\metricStationary}{Stationary}
\newcommand{\metricPopulation}{Population}
\newcommand{\metricFlow}{Flow}
\newcommand{\metricNNDist}{Nearest Neighbor Distance}
\newcommand{\metricMOTConf}{MOT Confidence}
\newcommand{\metricGeoConf}{3D Geometric Confidence}
\newcommand{\metricDisappearance}{Track Disappearance}

\newcommand{\mvelEMD}{\mathcal{M}_{\text{vel}}^{\text{E}}}
\newcommand{\mvel}{\mathcal{M}_{\text{vel}}}
\newcommand{\maccelEMD}{\mathcal{M}_{\text{acc}}^{\text{E}}}
\newcommand{\maccel}{\mathcal{M}_{\text{acc}}}
\newcommand{\mdistEMD}{\mathcal{M}_{\text{dist}}^{\text{E}}}
\newcommand{\mdist}{\mathcal{M}_{\text{dist}}}
\newcommand{\mdtw}{\mathcal{M}_{\text{path}}^{\text{DTW}}}
\newcommand{\mdiv}{\mathcal{M}_{\text{div}}^{\text{DTW}}}
\newcommand{\mintdiv}{\mathcal{M}_{\text{int-div}}^{\text{DTW}}}
\newcommand{\mcollEMD}{\mathcal{M}_{\text{coll}}^{\text{E}}}
\newcommand{\mcollrate}{\mathcal{M}_{\text{coll}}}
\newcommand{\mstatEMD}{\mathcal{M}_{\text{stat}}^{\text{E}}}
\newcommand{\mstat}{\mathcal{M}_{\text{stat}}}
\newcommand{\mpopEMD}{\mathcal{M}_{\text{pop}}^{\text{E}}}
\newcommand{\mpop}{\mathcal{M}_{\text{pop}}}
\newcommand{\mflowEMD}{\mathcal{M}_{\text{flow}}^{\text{E}}}
\newcommand{\mflow}{\mathcal{M}_{\text{flow}}}
\newcommand{\mnnEMD}{\mathcal{M}_{\text{nn}}^{\text{E}}}
\newcommand{\mnnmode}{\mathcal{M}_{\text{nn}}}
\newcommand{\mdisappear}{\mathcal{M}_{\text{disp}}}
\newcommand{\mdisappearEMD}{\mathcal{M}_{\text{disp}}^{\text{E}}}
\newcommand{\mmotconf}{\mathcal{M}_{\text{mot}}}
\newcommand{\mgeoconf}{\mathcal{M}_{\text{geo}}}

\section{Method}

Figure \ref{fig:method_details} summarizes \method. We generate videos under two conditions: I2V using start frames from real videos, and T2V using a structured prompt suite. Next, we extract pixel-space tracks with a multi-object tracker and convert them to metric bird's-eye view trajectories using dataset homographies for I2V or camera reconstruction plus metric depth and scale alignment for T2V. Finally, we quantify realism with a suite of kinematic, social interaction, and video-fidelity metrics.\\

\inlinesubsection{Problem Formulation.}
We consider a scene with a time-varying number of pedestrians (agents) depicted in a generated video $V^{\text{gen}}$ with $K$ frames. The state of the $i$-th agent at each time step $k \in \{0, \dots, K-1\}$ is its 2D bird's-eye view (BEV) position in world coordinates, $\mathbf{p}_k^i = (x_k^i, y_k^i) \in \mathbb{R}^2$. A trajectory is the time-ordered sequence of positions for a unique agent, $\mathcal{T}^i = (\mathbf{p}_k^i)_{k=k_{\text{start}}^i}^{k_{\text{end}}^i}$, active from its entry time step $k_{\text{start}}^i$ to its exit time step $k_{\text{end}}^i$. The length of each trajectory is $L_i = k_{\text{end}}^i - k_{\text{start}}^i + 1$. A complete crowd scene is the set of all such extracted trajectories, $\mathcal{X} = \{\mathcal{T}^1, \mathcal{T}^2, \dots, \mathcal{T}^{|\mathcal{X}|}\}$. The total number of unique trajectories, or scene cardinality, is $|\mathcal{X}|$. We use the shorthand $N_{\text{gen}} = |\mathcal{X}^{\text{gen}}|$ and $N_{\text{gt}} = |\mathcal{X}^{\text{GT}}|$ for generated and ground-truth scenes, respectively.\\

\inlinesubsection{Image Prompts.}
To enable direct comparison with ground-truth dynamics, we condition I2V models on start frames from the ETH \cite{Lerner_ETH} and UCY \cite{Pellegrini_UCY} pedestrian trajectory datasets. There are five scenes in total: ETH, HOTEL, UNIV, ZARA1, and ZARA2. From each of the scenes, we extract non-overlapping start frames at 5-second intervals, creating an image prompt suite of 530 unique frames. The images extracted from ETH/UCY are used as conditioning images. Each image is used to generate one video per model.
However, we repeat the generation process to gather sufficient pedestrian data from each scene, performing multiple inferences until accumulating at least $N_{\text{gen}}=150$ unique tracks or 1500 total detections (Table \ref{tab:dataset_comprehensive_stats} with details is provided in the Appendix). 
Each generation is conditioned on the image and a constant text prompt: ``\textit{A stationary overhead view of pedestrian movement.}''\\

\inlinesubsection{Text Prompts.}\label{sec:pedra_t2v}
We develop a structured prompt suite to systematically evaluate T2V models across diverse pedestrian scenarios. Prompts are organized along two axes: \textbf{crowd density} and pedestrian \textbf{interaction type}.
We define three levels for each axis:
\begin{itemize}
    \item Sparse (\textbf{\denS}): The scene contains very few people, often individuals or small, separated groups.
    \item Moderate (\textbf{\denM}): A comfortable number of people are present to make the area feel active.
    \item Crowded (\textbf{\denC}): The area is densely populated, and movement is visibly constrained by others.
\end{itemize}
We additionally define three \textbf{interaction types}:
\begin{itemize}
    \item Directional (\textbf{\intD}): The majority of pedestrians are moving in a clear, linear pattern along one dominant axis.
    \item Multidirectional (\textbf{\intM}): Pedestrians are moving in many different directions without a single dominant axis.
    \item Converging/Diverging (\textbf{\intC}): Pedestrian movement is oriented around a specific point of interest or bottleneck.
\end{itemize}
Using these definitions, we prompt an LLM (Gemini 2.5 Pro) to generate 20 distinct scene descriptions for each of the nine density/interaction categories, always requesting a stationary camera viewpoint. We sample 5 repetitions for each prompt, generating 900 videos (1.25 hours) per model.\\

\inlinesubsection{Trajectory Extraction for I2V.}
The initial step is to extract 2D pedestrian trajectories in pixel coordinates from the generated videos (Fig. \ref{fig:method_details}). We use FairMOT \cite{zhang_fairmot_2021} as a performant off-the-shelf multi-object tracker (MOT), though our framework permits the use of any equivalent MOT choice. The tracker processes each video $V^{\text{gen}}$ to produce a set of pixel-space tracklets $\{\mathcal{T}^i_{\text{px}}\}$. From this output, we estimate the ground contact point for each person in each frame as the bottom-midpoint of their corresponding bounding box. To ensure a fair comparison between synthetic and real-world videos, we reduce label bias by reprocessing the ground-truth videos with the same MOT pipeline instead of using the original manual annotations. We project pixel-space tracks into BEV world coordinates using the pre-computed homography matrices from the ETH/UCY datasets. 
Before processing, videos are resized to their original source resolution, and we filter for static camera viewpoints using pyramidal Lucas-Kanade optical flow \cite{lucas1981iterative,bouguet2001pyramidal}.\\

\inlinesubsection{3D Reconstruction and Scale Estimation for T2V.}
Reconstructing metric-scale trajectories from T2V outputs presents a significant challenge, as the generated scenes lack any known camera parameters, 3D geometry, or guaranteed static viewpoints \cite{he_cameractrl_2025,he2025cameractrliidynamicscene,ren2025gen3c}. To address this, we propose a pipeline to recover BEV trajectories (Fig. \ref{fig:method_details}a). We first use VGGT \cite{VGGT_Wang_2025_CVPR} to estimate per-frame camera intrinsics ($K_k$), extrinsics ($R_k, t_k$), and a geometrically consistent but unscaled depth map $D_{\text{norm}, k}$.

To establish a real-world scale, we follow He et al. \cite{he2025cameractrliidynamicscene} and employ a separate metric depth estimator, Depth Pro \cite{depthpro_bochkovskiy_2025}, on keyframes to generate metric-scale depth maps $D_{\text{metric}, k}$. 
We then compute frame-by-frame scale factors by robustly aligning $D_{\text{metric}, k}$ and $D_{\text{norm}, k}$ using a RANSAC (Random Sample Consensus) algorithm \cite{hartley2003multiple}. In each RANSAC iteration, we solve for the per-frame scale $\lambda_k$ by minimizing a Huber loss between the scaled VGGT depth and the metric depth \cite{he2025cameractrliidynamicscene}: 
\begin{equation}
\lambda_k = \argmin_{\lambda'} \sum_{p \in \mathcal{P}} \rho\left(|\lambda' \cdot D_{\text{norm}, k}(p) - D_{\text{metric}, k}(p)|\right)
\end{equation}
where $\mathcal{P}$ is the set of valid pixels and $\rho(\cdot)$ is the Huber loss.

As a final validation step, we enforce an anthropometric prior. We use the scaled camera parameters to estimate the real-world height of each detected person using the pinhole camera projection formula $H_{\text{world}} = h_{\text{pixels}} \cdot Z_{\text{cam}} / f_y$, where the depth $Z_{\text{cam}}$ is derived from our scaled 3D reconstruction, $h_{\text{pixels}}$ is the bounding box height, and $f_y$ is the camera's vertical focal length. If the mean height across all detections falls outside a plausible range of $(1.4, 2.0)$~meters \cite{owid-human-height}, we correct the scale factors $\lambda_k$ to align the mean height to $1.7$~m.
Finally, we apply the validated scale factor to the camera trajectory and un-project the MOT pixel tracks into a unified, meter-scale BEV coordinate system, yielding the final trajectory set $\mathcal{X}^{\text{gen}}$. We apply this correction only if the two depth map estimates are consistent apart from scale, and otherwise discard the video sample.

\section{Evaluation Metrics}
\label{sec:eval}

\begin{table*}[t]
    \centering
    \small
    \setlength{\tabcolsep}{3mm}
    \begin{tabular}{@{}ll@{}}
    \toprule
    \textbf{Category} & \textbf{Measure (Symbol)} \\
    \midrule
    Trajectory Kinematics (\hyperref[category:kinematics]{\S\ref{category:kinematics}}) & Velocity ($\mvel$), Acceleration ($\maccel$), Distance ($\mdist$) \\
    Social Interaction (\hyperref[category:social]{\S\ref{category:social}}) & Collision ($\mcollrate$), Stationary ($\mstat$), Population ($\mpop$), Flow ($\mflow$), NN Dist. ($\mnnmode$) \\
    Video Fidelity (\hyperref[category:fidelity]{\S\ref{category:fidelity}}) & Disappearance ($\mdisappear$), MOT Confidence ($\mmotconf$), 3D Geometric Confidence ($\mgeoconf$) \\
    \bottomrule
    \end{tabular}
    \caption{List of measures and their symbols. For I2V metrics, EMD-based comparisons to ground truth distributions are denoted with a superscript (e.g., $\mvelEMD$).}
    \label{tab:metrics_inline}
\end{table*}

We propose a suite of metrics to assess the realism of generated pedestrian dynamics, summarized in Table~\ref{tab:metrics_inline}. \emph{Complete mathematical definitions and implementations are provided in Appendix Section~\ref{sec:metricstheory}.}

Our evaluation protocol distinguishes between I2V and T2V tasks. For T2V, which lacks ground truth, we report measures in real-world units (e.g., meters, percentages) and compare against reference ranges computed from ten public pedestrian benchmarks: ETH~\cite{Lerner_ETH}, UCY~\cite{Pellegrini_UCY}, PETS-2009~\cite{PETS2009}, SDD~\cite{SDD_2016}, Grand Central~\cite{GC_dataset}, HERMES~\cite{HERMES_bottleneck}, KITTI~\cite{KITTI_2012}, Edinburgh~\cite{majecka2009statistical}, Town Center~\cite{towncenter}, and WildTrack~\cite{chavdarova2017wildtrack}; all accessed via OpenTraj~\cite{amirian2020opentraj}.
The benchmarks span a variety of settings, densities, and interaction types, forming a diverse \emph{real-world reference} for realistic ranges of the evaluation measures (henceforth labeled \emph{\textbf{``Ref.''}}).

For I2V, where ground truth video exists, we measure distributional dissimilarity using Earth Mover's Distance (EMD)~\cite{rubner_emd_1998}, which was recently proposed for crowd simulation evaluation~\cite{bae_continuous_2025}. 
EMD measures the minimal work required to transform one distribution into another. Given two discrete distributions $P=\{p_1, \ldots, p_m\}$ and $Q=\{q_1, \ldots, q_n\}$, EMD finds an optimal flow $F=\{f_{ij}\}$ that minimizes the total cost $\sum_{i,j} f_{ij} d_{ij}$, where $d_{ij}$ is the distance between elements $p_i$ and $q_j$. 
Since the metrics consider the entire set of generated trajectories, the I2V metrics do not expect deterministic replication of particular GT trajectories; rather, lower EMD indicates higher statistical agreement between the overall metric distributions in the simulation and GT.
We normalize all EMD computations using ground truth distribution parameters ($\mu^{\text{GT}}$, $s^{\text{GT}}$):
\begin{equation}
    \textstyle \emdsymb(\mathcal{A}, \mathcal{B}) \coloneqq \text{EMD}\!\left(\frac{\mathcal{A}-\mu^{\text{GT}}}{s^{\text{GT}}}, \frac{\mathcal{B}-\mu^{\text{GT}}}{s^{\text{GT}}}\right)
\end{equation}
where $\mathcal{A}$ and $\mathcal{B}$ represent arbitrary sample sets from the generated and ground truth data, respectively.

\subsection{Trajectory Kinematics}\label{category:kinematics}
These metrics assess physical plausibility of individual agent movements, capturing how pedestrians walk and change direction.\\

\inlinesubsection{\metricVelocity.}\label{metric:velocity} 
Pedestrian walking speed is fundamental to human locomotion with well-documented distributions and physically plausible limits.
We use a Kalman smoother to estimate velocities and compute the per-agent average speed $\bar{s}^i$ for each agent $i$. 
For T2V, the mean speed across all $N_{\text{gen}}$ generated agents is computed as $\textstyle \mvel = \frac{1}{N_{\text{gen}}} \sum_{i=1}^{N_{\text{gen}}} \bar{s}^i$.
For I2V, we compare the distribution of per-agent average speeds between generated and ground truth scenes using $\mvelEMD = \emdsymb(\{\bar{s}^i\}_{\text{gen}}, \{\bar{s}^j\}_{\text{GT}})$, where the subscripts denote the set of per-agent average speeds from generated and ground truth trajectories.\\

\inlinesubsection{\metricAcceleration.}\label{metric:acceleration} 
When people walk, they can't instantly speed up or turn. Walking acceleration provides a measure of whether the simulated motion obeys realistic human biomechanical constraints. Abnormally high accelerations would be problematic. We compute the per-agent average acceleration magnitude $\bar{a}^i$ for each agent $i$. 
For T2V, the mean acceleration is $\textstyle\maccel = \frac{1}{N_{\text{gen}}} \sum_{i=1}^{N_{\text{gen}}} \bar{a}^i$.
For I2V, we compute ${\maccelEMD = \emdsymb(\{\bar{a}^i\}_{\text{gen}}, \{\bar{a}^j\}_{\text{GT}})}$.\\

\inlinesubsection{\metricDistance\ Traveled.}\label{metric:distance} 
Total distance traversed reflects pedestrians' purposefulness and environmental engagement. Agents moving too little or covering implausible distances fail to capture realistic behavior. We measure the total path length $d^i$ for each agent $i$. 
For T2V, the mean distance is $\textstyle\mdist = \frac{1}{N_{\text{gen}}} \sum_{i=1}^{N_{\text{gen}}} d^i$.
For I2V, we compute ${\mdistEMD = \emdsymb(\{d^i\}_{\text{gen}}, \{d^j\}_{\text{GT}})}$.

\begin{figure*}[t!]
    \centering
    \includegraphics[width=\linewidth]{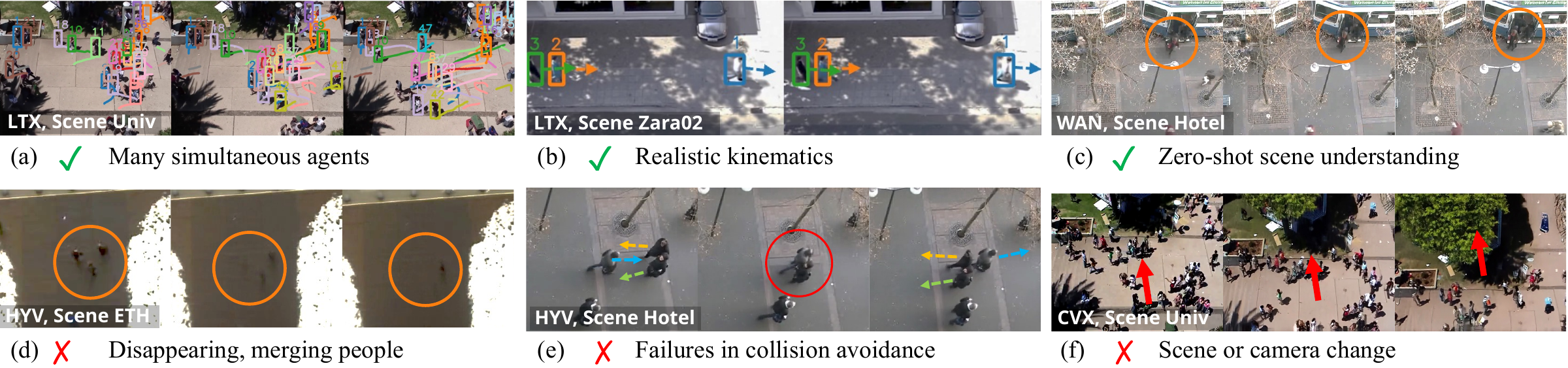}
    \caption{Examples of common successes and failures illustrated through the I2V benchmark.}
    \label{fig:qual_i2v}
\end{figure*}

\subsection{Social Interaction}\label{category:social}

These metrics evaluate realistic multi-agent behaviors. Human crowds exhibit complex emergent phenomena including collision avoidance, personal space preferences, density-dependent speed regulation, and collective patterns.\\

\inlinesubsection{\metricCollision\ Rate.}\label{metric:collision} 
Real pedestrians actively avoid collisions through anticipatory behaviors. High collision rates indicate failure to model spatial reasoning and social interaction. We define a collision as occurring when agents are within $\delta=0.1$ meters of each other. Let $\mathbb{I}_{\text{coll}}(i,k)$ be an indicator function that equals 1 if agent $i$ is in collision at time $k$, and 0 otherwise.
For T2V, the collision rate $\mcollrate$ is the percentage of all detected agents across all frames that are in a collision state: $\textstyle\mcollrate = \frac{100}{N_{\text{total}}} \sum_{\mathcal{T}^i \in \mathcal{X}^{\text{gen}}} \sum_{k=k_{\text{start}}^i}^{k_{\text{end}}^i} \mathbb{I}_{\text{coll}}(i,k)$, where $N_{\text{total}} = \sum_{\mathcal{T}^i \in \mathcal{X}^{\text{gen}}} L_i$ represents the total count of all agent detections over all frames.
For I2V, we compute the per-frame collision count $N_{\text{coll}}(k) = \sum_{i \in \mathcal{A}_k} \mathbb{I}_{\text{coll}}(i,k)$ for each frame $k$, where $\mathcal{A}_k$ is the set of active agents at time $k$. The metric $\mcollEMD = \emdsymb(\{N_{\text{coll}}(k)\}_{\text{gen}}, \{N_{\text{coll}}(k)\}_{\text{GT}})$ compares the distribution of per-frame collision counts.\\

\inlinesubsection{\metricStationary\ Agents.}\label{metric:stationary} 
Public contain people who stop to talk, wait, or sit. This social behavior is common but often overlooked by trajectory models focused on locomotion or per-agent marginal predictions. The \method\ T2V benchmark specifically includes public scenes that should include some seated and standing people.
An agent $i$ is classified as stationary ($\mathbb{I}_{\text{stat}}(i) = 1$) if its end-to-end displacement is less than $0.2$ meters. 
For T2V, the percentage of stationary agents is $\textstyle\mstat = \frac{100}{N_{\text{gen}}} \sum_{i=1}^{N_{\text{gen}}} \mathbb{I}_{\text{stat}}(i)$.
For I2V, we compute $\mstatEMD = \emdsymb(\{\mathbb{I}_{\text{stat}}(i)\}_{\text{gen}}, \{\mathbb{I}_{\text{stat}}(j)\}_{\text{GT}})$.\\

\inlinesubsection{\metricPopulation.}\label{metric:population} 
The population in the video frame reflects the density of the scene.
For T2V, this metric tests whether models respond to density prompts (``crowded'' vs.\ ``sparse''). For I2V, it evaluates replication of populations that would be typical in the scene provided to the model. The population at time $k$ is defined as the number of active agents, denoted $|\mathcal{A}_k|$. 
For T2V, the mean population (average number of agents per frame) is $\textstyle\mpop = \frac{1}{K} \sum_{k=0}^{K-1} |\mathcal{A}_k^{\text{gen}}|$, where $K$ is the total number of frames. 
For I2V, we compute $\mpopEMD = \emdsymb(\{|\mathcal{A}_k|\}_{\text{gen}}, \{|\mathcal{A}_k|\}_{\text{GT}})$.\\

\inlinesubsection{\metricFlow.}\label{metric:flow} 
A well-known principle in pedestrian dynamics is the fundamental diagram: as local density increases, pedestrians slow due to spatial constraints and social forces \cite{Seyfried_2005}. Models that violate this inverse density-speed relationship generate unrealistic crowds, such as maintaining high walking speeds when packed. For agent $i$ at time $k$, we compute local density $\rho_k^i$ (agents/m$^2$) using the area of the circle enclosing its $K=4$ nearest neighbors, and instantaneous flow $f_k^i = \rho_k^i \cdot \| \mathbf{v}_k^i \|_2$. For T2V, the mean flow averages the x and y directional flows, $\textstyle\mflow = \frac{1}{2} (\bar{f}_x + \bar{f}_y)$. For I2V, the metric is $\textstyle\mflowEMD = \frac{1}{2} \left( \emdsymb(\mathcal{F}_x^{\text{gen}}, \mathcal{F}_x^{\text{GT}}) + \emdsymb(\mathcal{F}_y^{\text{gen}}, \mathcal{F}_y^{\text{GT}}) \right)$ where $\mathcal{F}_x$, $\mathcal{F}_y$ are flow distributions partitioned by primary movement direction.\\

\inlinesubsection{\metricNNDist}\label{metric:nndist} 
Humans maintain characteristic personal space distances varying with context but following consistent patterns.
Interestingly, the nearest neighbor (NN) distance typically follows a bimodal distribution with peaks at distances around 0.5-0.75 meters \cite{minartz_necs_2025}.
For each moving agent (speed $> 0.1$ m/s), we find the Euclidean distance away to its nearest moving neighbor.
Let $D_{\text{nn}}$ denote the set of all NN distances. 
For T2V, we report the mean NN distance: $\mnnmode = \frac{1}{|D_{\text{nn}}^{\text{gen}}|} \sum_{d \in D_{\text{nn}}^{\text{gen}}} d$. 
For I2V, we compute the normalized EMD $\mnnEMD = \emdsymb(D_{\text{nn}}^{\text{gen}}, D_{\text{nn}}^{\text{GT}})$.

\begin{figure*}[t]
    \centering
    \includegraphics[width=\linewidth]{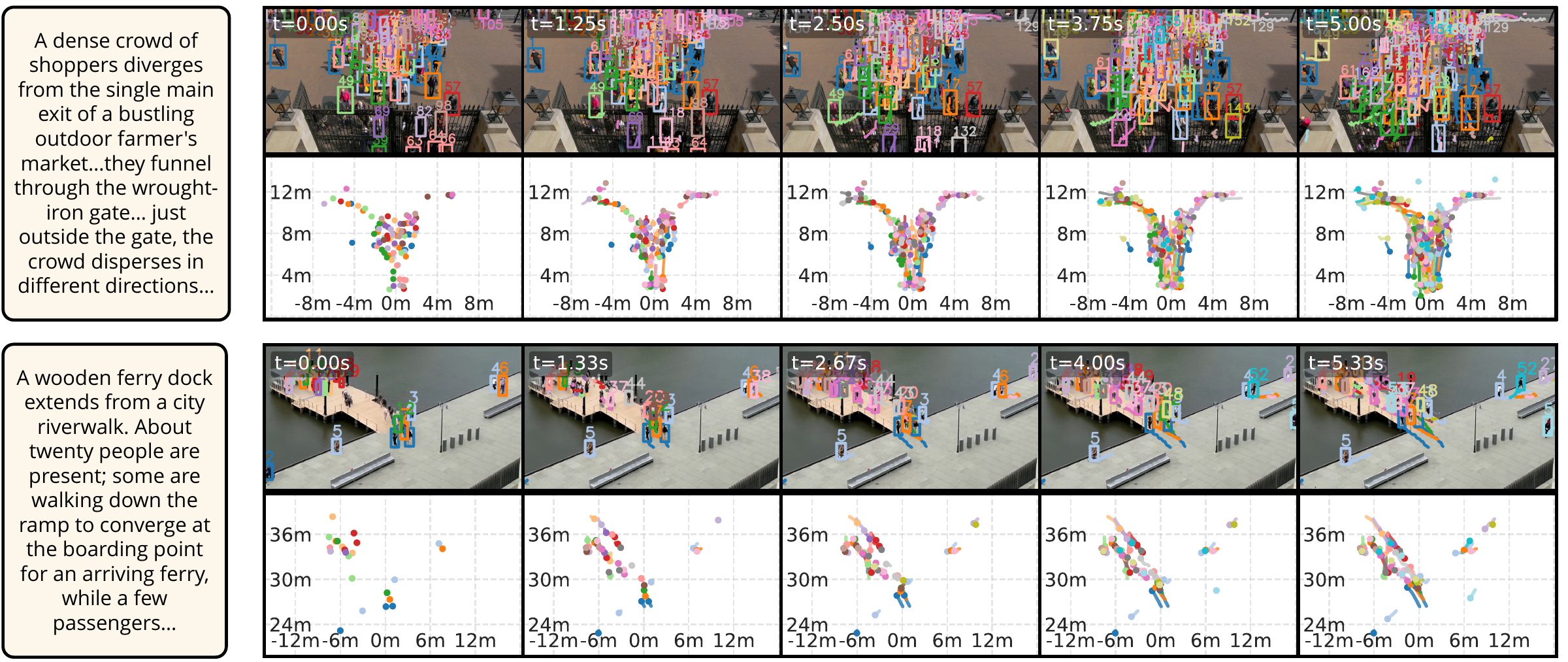}
    \caption{T2V results highlight the models' ability to generate complex social behaviors and scenes from text prompts. Additional visualizations provided in Appendix Figure \ref{fig:qualitative_examples}.}
    \label{fig:qual_t2v}
\end{figure*}

\subsection{Video Fidelity}\label{category:fidelity}
These metrics measure video quality and tracking reliability. Poor visual quality or geometric inconsistencies undermine utility for real-world applications like training perception systems.\\

\inlinesubsection{\metricDisappearance.}\label{metric:disappearance}
This metric measures how often trackers lose the trajectories of identified individuals.
It reports the proportion of MOT tracks that terminate within the central area of the frame, excluding a margin on each edge.
Mid-frame track terminations can reflect tracking failures, occlusion, or visual degradation where pedestrians become undetectable or vanish. Elevated disappearance rates compared to real-world videos indicate quality problems.
An agent $i$ is classified as disappearing mid-scene ($\mathbb{I}_{\text{disp}}(i) = 1$) if its final detected position lies within the central 80\% of the frame (defined by excluding a 10\% margin on all edges). 
For T2V, the mid-scene disappearance rate is $\textstyle\mdisappear = \frac{100}{N_{\text{total}}} \sum_{i=1}^{N_{\text{total}}} \mathbb{I}_{\text{disp}}(i)$, where $N_{\text{total}}$ is the total number of unique agents in the generated scene. 
For I2V, we compute $\mdisappearEMD = \emdsymb(\{\mathbb{I}_{\text{disp}}(i)\}_{\text{gen}}, \{\mathbb{I}_{\text{disp}}(j)\}_{\text{GT}})$.\\

\inlinesubsection{\metricMOTConf.}\label{metric:motconf}
This metric uses the confidence score from a pre-trained multi-object tracker \cite{zhang_fairmot_2021} as a proxy for the visual quality and trackability of generated pedestrians. 
Let $\sigma_k^i$ denote the tracker confidence score for agent $i$ at time $k$. The metric is computed as the mean of the per-agent average confidence scores: $\textstyle\mmotconf = \frac{1}{N_{\text{gen}}} \sum_{i=1}^{N_{\text{gen}}} \left( \frac{1}{L_i} \sum_{k=k_{\text{start}}^i}^{k_{\text{end}}^i} \sigma_k^i \right)$, where $L_i$ is the length of agent $i$'s trajectory. This metric is used for both I2V and T2V evaluations, and higher is better.

\vspace{.5em}
\inlinesubsection{\metricGeoConf.}\label{metric:geoconf}
We observe that synthetic videos with higher 3D reconstruction success have higher aesthetic quality overall.
We assess the 3D consistency using confidence scores from VGGT \cite{VGGT_Wang_2025_CVPR}. These scores are derived from predicted aleatoric uncertainty in depth estimation. Higher confidence indicates internally consistent geometry. Let $\gamma_k^i$ denote the confidence at agent $i$'s location at time $k$ (values are greater than 1, with values approaching 1 indicating high uncertainty). The metric is $\textstyle\mgeoconf = \frac{1}{N_{\text{gen}}} \sum_{i=1}^{N_{\text{gen}}} \left( \frac{1}{L_i} \sum_{k=k_{\text{start}}^i}^{k_{\text{end}}^i} \gamma_k^i \right)$.
This metric is used for both I2V and T2V evaluations.
\section{Experiments}
\newcommand{\WAN}{WAN}
\newcommand{\CVX}{CVX}
\newcommand{\HYV}{HYV}
\newcommand{\LTX}{LTX}
\newcommand{\OS}{OS}

We selected five SOTA models that have both I2V and T2V variants: Wan2.1 \cite{wan_wan_2025}, CogVideoX1.5 \cite{yang_cogvideox_2025}, HunyuanVideo \cite{kong_hunyuanvideo_2025}, LTX-Video \cite{HaCohen2024LTXVideo}, and Open-Sora 2.0 \cite{peng2025opensora20}. We refer to them as \WAN, \CVX, \HYV, \LTX, and \OS, respectively.
We standardize all generations to a $\sim$5-second duration, which is the maximum for \OS\ and HYV, with resolution as close as possible to the start image resolution for I2V and 720p for T2V. We include a typical negative prompt to discourage visual artifacts and camera motion. All models are run with their suggested default hyperparameters on four NVIDIA H200 GPUs, resulting in generation times varying between 2 and 8 minutes per video.
\emph{Full configuration details are provided in Appendix Section \ref{sec:inference_details}}.

\subsection{Qualitative Results}

\begin{table}[t]
    \centering
    \footnotesize
    \setlength{\tabcolsep}{3pt}
    \caption{I2V evaluation metrics averaged over the five ETH/UCY scenes. Lower values are better for all EMD metrics. Higher values are better for MOT Confidence. The best and second-best results in each column are in bold and underlined, respectively.}
    \resizebox{\columnwidth}{!}{%
    \label{tab:i2v_results}
    \begin{tabular}{c@{~~}c c c@{~}c c c c c@{~}c c}
    \toprule
    \multirow{2}{*}{Model} & \multicolumn{3}{c}{Trajectory Kinematics} & \multicolumn{5}{c}{Social Interaction} & \multicolumn{2}{c}{Fidelity} \\ \cmidrule(lr){2-4} \cmidrule(lr){5-9} \cmidrule(lr){10-11}
                           & $\mvelEMD$ & $\maccelEMD$ & $\mdistEMD$ & $\mcollEMD$ & $\mstatEMD$ & $\mpopEMD$ & $\mnnEMD$ & $\mflowEMD$ & $\mdisappearEMD$ & $\mmotconf$ \\ \midrule
        WAN             & \underline{0.457} & 0.782 & \underline{0.370} & \textbf{0.029} & \underline{0.159} & 0.701 & \underline{0.084} & 0.384 & 0.331 & 0.497 \\
        HYV             & \textbf{0.419} & \textbf{0.639} & \textbf{0.288} & \underline{0.033} & \textbf{0.135} & \textbf{0.467} & 0.100 & \underline{0.228} & \underline{0.158} & \underline{0.500} \\
        OS              & 0.549 & \underline{0.703} & 0.462 & 0.047 & 0.316 & 0.916 & 0.155 & \textbf{0.180} & 0.514 & 0.486 \\
        LTX             & 0.510 & 0.747 & 0.391 & 0.041 & 0.263 & \underline{0.568} & \textbf{0.051} & 0.745 & \textbf{0.130} & \textbf{0.503} \\
        CVX             & 0.808 & 0.706 & 0.621 & 0.054 & 0.280 & 0.892 & 0.137 & 0.228 & 0.169 & 0.491 \\
    \bottomrule
    \end{tabular}%
    }
\end{table}

The video models can generate multi-agent pedestrian scenes (Fig. \ref{fig:qual_i2v}a) with enough visual consistency in the trackable agents to extract trajectories. Many videos appear visually realistic because pedestrians move at natural-looking speeds (Fig.~\ref{fig:qual_i2v}b), though motion can be overly smooth or jittery. In the I2V task, generated pedestrians largely adhere to environmental constraints such as sidewalks. The models sometimes demonstrate surprising zero-shot scene understanding such as an arriving train opening its doors to pedestrians entering (Fig. \ref{fig:qual_i2v}c). For T2V, models successfully translate semantic prompts into intuitive social behaviors (Fig.~\ref{fig:qual_t2v}), for instance producing a converging funnel when a ``crowd of shoppers'' exits a market, or plausible walking paths on a pier that avoid the water and furniture. These outcomes are notable given the domain expertise and engineering required to produce similar results with conventional crowd simulators.

We also observe recurring failure modes that degrade physical and social plausibility. The primary issue is a lack of agent-level integrity: pedestrians sometimes merge into one another or disappear mid-trajectory (Fig.~\ref{fig:qual_i2v}d). Merging is especially common in T2V outputs for the ``crowded'' (\textbf{\denC}) and ``multidirectional'' (\textbf{\intM}) prompts, where models may produce untrackable, fluid-like pixel masses. Some prompts cause time-lapse effects that blur agents into streaks, examples of which appear in Appendix Fig.~\ref{fig:t2v_failure_modes}. People sometimes walk directly through each other as if the other approaching agents are completely absent (Fig. \ref{fig:qual_i2v}e). These artifacts are strongest for background agents that occupy few pixels, suggesting a link between representation scale and dynamic consistency.
Some models produce distorted objects or false-positive person detections. We also see failures in prompt or scene adherence, for example models ignoring negative prompts intended to keep the camera static (Fig \ref{fig:qual_i2v}f), resulting in unwanted camera motion. In other cases, models may misinterpret the scene context, such as by animating a parked car in a pedestrian zone where it should remain stationary. Models differ in how they populate scenes -- HunyuanVideo \cite{kong_hunyuanvideo_2025} often reduces agent counts over the 5-second clip, as many pedestrians vanish and few new ones appear. These limitations point to clear opportunities for improving video-based world simulation.

\subsection{Quantitative Results}

\begin{table}[t]
    \centering
    \small
    \setlength{\tabcolsep}{.5mm}
    \caption{T2V measures averaged across models, grouped by density and interaction categories (see \S\ref{sec:pedra_t2v}).}
    \label{tab:t2v_pivot_categories}
    \resizebox{\columnwidth}{!}{%
    \begin{tabular}{@{~~~}c@{~~~}c@{~~~}ccccccc}
    \toprule
    \multirow{2}{*}{Experiment} & \multirow{2}{*}{Category} & $\mvel$ & $\mcollrate$ & $\mpop$ & $\mflow$ & $\mnnmode$ & $\mmotconf$ & $\mgeoconf$ \\
    & & (m/s) & (\%) & (Count) & (1/m/s) & (m) &  &  \\ \midrule
    \textbf{Ref.} & \textbf{-} & \textbf{0.91} & \textbf{1.19} & \textbf{13.77} & \textbf{0.54} & \textbf{1.18} & \textbf{-} & \textbf{-}  \\
    \cmidrule{1-9}
    \multirow{3}{*}{Density} & \denS & 0.66 & 1.71 & 4.79 & 0.45 & 1.62 & 0.62 & 3.33 \\
     & \denM & 0.67 & 3.29 & 24.33 & 0.44 & 0.97 & 0.58 & 2.80 \\
     & \denC & 0.51 & 7.57 & 74.75 & 1.18 & 0.69 & 0.53 & 2.49 \\
    \cmidrule{1-9}
    \multirow{3}{*}{Interaction} & \intD & 0.66 & 7.14 & 32.56 & 1.64 & 0.73 & 0.55 & 2.10 \\
     & \intM & 0.55 & 4.00 & 41.06 & 0.52 & 0.93 & 0.54 & 2.79 \\
     & \intC & 0.47 & 8.19 & 32.59 & 0.90 & 0.73 & 0.55 & 2.68 \\
\bottomrule
    \end{tabular}%
    }
\end{table}

\begin{table*}[t]
    \centering
    \small
    \caption{Mean $\pm$ one standard deviation of the T2V evaluation measures for the five tested models, inclusive of all Density and Interaction categories. The real-world reference (\textbf{Ref.}) shows the same measures computed on a combined dataset of 10 public pedestrian benchmarks Note: Ref. does not include the Fidelity metrics (\S\ref{category:fidelity}) as they relate to video post-processing not conducted on the reference data.}
    \label{tab:t2v_evaluation_overall}
    \resizebox{\linewidth}{!}{%
    \begin{tabular}{@{~~}c@{~~~}c@{~~~}c@{~~~}c@{~~~}c@{~~~}c@{~~~}c@{~~~}c@{~~~}c@{~~~}c@{~~~}c@{~~~}c@{~~~}c@{~~~}c}
    \toprule
    \multirow{3}{*}{Model} & \multicolumn{3}{c}{Trajectory Kinematics} & \multicolumn{5}{c}{Social Interaction} & \multicolumn{3}{c}{Video Fidelity} \\
    \cmidrule(lr){2-4} \cmidrule(lr){5-9} \cmidrule(lr){10-12}
    & $\mvel$ & $\maccel$ & $\mdist$ & $\mcollrate$ & $\mstat$ & $\mpop$ & $\mflow$ & $\mnnmode$ & $\mdisappear$ & $\mmotconf$ $\uparrow$ & $\mgeoconf$ $\uparrow$ \\
    & (m/s) & (m/s$^2$) & (m) & (\%) & (\%) & (Count) & (1/m/s) & (m) & (\%) &  &  \\ \midrule
    \textbf{Ref.} & \textbf{0.91 $\pm$ 0.84} & \textbf{0.65 $\pm$ 0.98} & \textbf{3.62 $\pm$ 3.56} & \textbf{1.19 $\pm$ 11.28} & \textbf{0.19 $\pm$ 0.39} & \textbf{13.77 $\pm$ 19.07} & \textbf{0.54 $\pm$ 0.45} & \textbf{1.18 $\pm$ 1.51} & - & - & - \\
    \midrule
    WAN & 0.56 $\pm$ 0.86 & 0.83 $\pm$ 1.05 & 2.16 $\pm$ 3.34 & 5.33 $\pm$ 8.16 & 0.22 $\pm$ 0.41 & 56.83 $\pm$ 79.76 & 1.30 $\pm$ 12.76 & 0.72 $\pm$ 0.95 & 29.19 $\pm$ 48.29 & 0.54 $\pm$ 0.11 & 3.02 $\pm$ 4.81 \\
    HYV & 0.66 $\pm$ 0.98 & 0.91 $\pm$ 1.08 & 1.56 $\pm$ 2.23 & 9.88 $\pm$ 13.31 & 0.25 $\pm$ 0.44 & 35.36 $\pm$ 46.65 & 1.73 $\pm$ 9.79 & 0.69 $\pm$ 1.03 & 29.11 $\pm$ 49.99 & 0.54 $\pm$ 0.11 & 2.13 $\pm$ 2.63 \\
    OS & 0.38 $\pm$ 0.59 & 0.50 $\pm$ 0.80 & 1.62 $\pm$ 2.02 & 2.68 $\pm$ 5.36 & 0.28 $\pm$ 0.45 & 22.79 $\pm$ 22.28 & 0.21 $\pm$ 0.46 & 1.06 $\pm$ 1.23 & 31.89 $\pm$ 46.59 & 0.56 $\pm$ 0.12 & 3.35 $\pm$ 5.15 \\
    LTX & 0.80 $\pm$ 1.06 & 1.19 $\pm$ 1.15 & 2.03 $\pm$ 2.73 & 7.62 $\pm$ 8.52 & 0.24 $\pm$ 0.43 & 33.76 $\pm$ 44.18 & 1.14 $\pm$ 3.06 & 0.74 $\pm$ 0.97 & 32.19 $\pm$ 48.40 & 0.56 $\pm$ 0.12 & 1.45 $\pm$ 1.37 \\
    CVX & 0.40 $\pm$ 0.59 & 0.65 $\pm$ 0.91 & 1.20 $\pm$ 1.78 & 5.77 $\pm$ 9.10 & 0.31 $\pm$ 0.46 & 28.51 $\pm$ 33.03 & 0.50 $\pm$ 1.05 & 0.80 $\pm$ 1.01 & 33.56 $\pm$ 49.41 & 0.51 $\pm$ 0.09 & 2.82 $\pm$ 3.45 \\
\bottomrule
    \end{tabular}%
    }
\end{table*}

\inlinesubsection{I2V.}
The benchmarking results are provided in Table \ref{tab:i2v_results}.
The results reveal that no single model consistently outperforms others across all metrics. HunyuanVideo \cite{kong_hunyuanvideo_2025} excels in trajectory kinematics with the best scores in the velocity, acceleration, and distance metrics.
LTX-Video \cite{HaCohen2024LTXVideo} produces the most visually coherent and trackable pedestrians as judged by its top scores in the $\mmotconf$ and $\mdisappearEMD$ metrics. Open-Sora \cite{peng2025opensora20} best recreated realistic pedestrian flow volume, and Wan \cite{wan_wan_2025} best recreated the low collision rate present in the ground truth video.\\

\inlinesubsection{T2V.}
Table \ref{tab:t2v_pivot_categories} shows that, on average, the models correctly interpret semantic prompts for crowd density and interaction type. They generate larger populations ($\mpop$) for ``crowded'' (\denC) compared to ``sparse'' (\denS) prompts and higher mean velocities ($\mvel$) for ``directional'' than for ``multidirectional'' or ``converging'' prompts. Collision rates ($\mcollrate$) also rise with density; for ``crowded'' scenes, more than 7\% of pedestrians are in contact with another agent in a given frame. While partially reflecting real-world crowding, this also suggests failures in collision avoidance. Still, the consistent response to semantic inputs suggest that models have learned latent mappings between text descriptions of density or interaction and their visual realization.

Compared to real-world data (Table~\ref{tab:t2v_evaluation_overall}), most models underestimate pedestrian speeds, with LTX-Video~\cite{HaCohen2024LTXVideo} coming closest, though it shows unrealistically high accelerations. Open-Sora~\cite{peng2025opensora20} achieves the lowest collision rate among T2V models, yet still exceeds real-world levels by more than twofold, and best matches plausible nearest-neighbor distances. All models overestimate population counts. Although the \textbf{Ref.} data includes dense scenes such as HERMES~\cite{HERMES_bottleneck}, it likely represents lower average densities than \method's ``crowded'' wide-angle scenes. HunyuanVideo~\cite{kong_hunyuanvideo_2025} exhibits a mean flow rate over three times the reference, exceeding physically realistic limits.\\

\inlinesubsection{Impact of Model-Specific Characteristics.}
Model performance depends on both architecture and training data. All models in this study use the DiT backbone, but they differ in VAE compression. LTX-Video \cite{HaCohen2024LTXVideo} and Open-Sora \cite{peng2025opensora20} apply high spatial compression for efficiency, which may blur individual agents in dense crowds. In contrast, Wan \cite{wan_wan_2025}, CogVideoX \cite{yang_cogvideox_2025}, and HunyuanVideo \cite{kong_hunyuanvideo_2025} use moderate compression that better preserves detail at higher cost. This did not benefit CogVideoX, which still showed high distortion and inconsistency in both I2V and T2V. Training data also play a major role. Some models employ auxiliary post-training such as spatial relation learning (Wan) or dense captioning (CogVideoX). Although models are exposed to pedestrians in web-scale datasets, this may be counteracted by data filtering choices. For instance, WAN explicitly removes ``crowded street scenes'' \cite{wan_wan_2025} to improve motion clarity, and HunyuanVideo filters out videos with more than five people during fine-tuning for certain downstream tasks \cite{kong_hunyuanvideo_2025}. Such choices favor single-subject realism over complex multi-agent dynamics, which might explain worse quality in dense scenarios.

\inlinesubsection{Limitations.}
Our benchmark has two primary limitations. First, our multi-stage trajectory extraction pipeline can introduce label noise, particularly in T2V metric scale estimation. We mitigate this by checking that human height falls within feasible ranges and filtering low-confidence outputs.
Second, our scope is limited to a representative set of current models and their short (5-second) generation horizon. This precludes an analysis of long-range navigation or how simulation fidelity degrades over time, which are aspects captured by traditional crowd simulation.

\section{Conclusion}
In this work, we introduce a new paradigm for evaluating video generation models as implicit simulators of complex multi-person behavior. We propose \method, a benchmark protocol that assesses the physical and social realism of pedestrian dynamics using a novel method to extract 2D trajectories from synthetic videos.
Our analysis reveals that leading models have learned an effective prior for plausible multi-agent behavior, successfully translating high-level prompts about crowd density and interaction into coherent motion and even emergent social phenomena. However, this success is frequently undermined by consistent failure modes, such as pedestrians merging or spontaneously disappearing, which pinpoint key areas for improvement. By establishing a rigorous evaluation framework and a public dataset, this work provides a foundation for developing next-generation world models capable of simulating the dynamics of human interaction in shared spaces.
\section*{Acknowledgments}
Thank you to Minartz et al. (\href{https://github.com/kminartz/NeCS}{NeCS}) and Bae et al. (\href{https://github.com/InhwanBae/Crowd-Behavior-Generation}{CrowdES}) for their code, which served as a base for some of the metrics and plots. 

Financial support for the authors was provided by the U.S. ONR under grant N00014-23-1-2799.
{
    \small
    \bibliographystyle{ieeenat_fullname}
    \bibliography{references}
}


\clearpage
\maketitlesupplementary

\appendix
\renewcommand{\thesection}{\Alph{section}}
\renewcommand{\thesubsection}{\thesection.\arabic{subsection}}

\section*{Appendix}

This section includes additional background, details on the methodology, and extended results and analysis. We first provide a brief summary of \textbf{video diffusion models} (\S\ref{sec:appdx_related}). The \textbf{Evaluation Metrics} (\S\ref{sec:metricstheory}) section covers implementation details not provided in the main paper. We describe the \textbf{prompt suite} (\S\ref{sec:prompt_suite}) and \textbf{inference process} (\S\ref{sec:inference_details}) including computing hardware and hyperparameters. We provide the additional image-to-video and text-to-video \textbf{results plots} (\S\ref{sec:quantitative_results}). Finally, we show additional \textbf{qualitative examples} (\S\ref{sec:qualitative_results}) of the video generations and postprocessing results, including successes and common failure modes.

\section{Background on Video Diffusion Models}
\label{sec:appdx_related}
Modern video diffusion models (VDMs) use the latent diffusion paradigm \cite{ho_ddpm_2020, rombach2022high}, which performs denoising in a compressed variational autoencoder (VAE) latent space. Early approaches adapted 2D U-Net \cite{ronneberger2015unet} backbones by either inserting temporal modules into frozen spatial layers \cite{ho2022videodiffusionmodels, blattmann_align_2023, wang_lavie_2025, chen_videocrafter2_2024} or using unified space-time architectures \cite{bartal2024lumiere}.
A subsequent architectural shift replaced the U-Net with the more scalable Diffusion Transformer (DiT) backbone \cite{peebles2023scalable, esser2024mmdit}, which now underpins many SOTA models \cite{chen2023pixartalpha, wan_wan_2025, kong_hunyuanvideo_2025, yang_cogvideox_2025, nvidia2025cosmos, polyak2025moviegencastmedia, gao2025seedance10}.
Conditioning signals are typically integrated via cross-attention \cite{chen2023pixartalpha}, adaptive layer normalization \cite{peebles2023scalable}, or unified self-attention across modalities \cite{ju2025fulldit, ren2025gen3c}.
In standard DiT training for video models, there is no explicit object or agent representation or any type of multi-agent inductive bias.

\section{Evaluation Metrics}
\label{sec:metricstheory}

Our evaluation protocol distinguishes between the I2V and T2V tasks as described in the main text.
We largely draw inspiration from two recent sources. From Bae et al. \cite{bae_continuous_2025} we adapt \metricVelocity, \metricAcceleration, and \metricDistance, and \metricPopulation. The primary change is that we compute per-agent rather than per-frame averages to allow for multiple repetitions of the same simulation. From Minartz et al. \cite{minartz_necs_2025} we adapt Nearest Neighbor Distance and \metricFlow.

\subsection{Trajectory Kinematics}

We first employ a Kalman smoother, $\mathcal{K}$, to estimate smoothed position $\tilde{\mathbf{p}}_k^i$ and velocity $\mathbf{v}_k^i$ states. For each agent $i$, we compute $(\tilde{\mathbf{p}}_k^i, \mathbf{v}_k^i)_{k=k_{\text{start}}^i}^{k_{\text{end}}^i} = \mathcal{K}(\mathcal{T}^i)$.
Using the estimated velocity, we compute the average speed $\bar{s}^i$ for each agent $i$ over its trajectory:
\begin{equation*}
    \bar{s}^i = \frac{1}{L_i-1} \sum_{k=k_{\text{start}}^i+1}^{k_{\text{end}}^i} \| \mathbf{v}_k^i \|_2
\end{equation*}

From the velocities $\mathbf{v}_k^i$, we compute the instantaneous acceleration using finite difference of velocity, then compute a per-agent average acceleration magnitude:
\begin{equation*}
    \bar{a}^i = \frac{1}{L_i - 2} \sum_{k=k_{\text{start}}^i+2}^{k_{\text{end}}^i} \frac{\| \mathbf{v}_k^i - \mathbf{v}_{k-1}^i \|_2}{t_k - t_{k-1}}
\end{equation*}
where $t_k$ is the timestamp (in seconds) for frame $k$.

The total path length $d^i$ is calculated by summing the Euclidean distance between consecutive points along the smoothed agent path:
\begin{equation*}
    d^i = \sum_{k=k_{\text{start}}^i+1}^{k_{\text{end}}^i} \| \tilde{\mathbf{p}}_k^i - \tilde{\mathbf{p}}_{k-1}^i \|_2
\end{equation*}

\subsection{Social Interaction}

\inlinesubsection{\metricCollision\ Rate.}
We define a collision for an agent as any instance where another agent is within a distance threshold $\delta=0.1$ meters. Define the set of agent indices active at a specific time step $k$ as $\mathcal{A}_k$.
The indicator function $\mathbb{I}_{\text{coll}}(i,k)$ is 1 if agent $i$ is in a collision at time $k$, and 0 otherwise:
\begin{equation*}
\mathbb{I}_{\text{coll}}(i,k) = 1 \iff \exists j \in \mathcal{A}_k, j \neq i : \| \mathbf{p}_k^i - \mathbf{p}_k^j \|_2 < \delta
\end{equation*}
These indicators are summed over the full set of generated trajectories to compute $\mcollrate$ and $\mcollEMD$ (see Section~\ref{metric:collision}).

\inlinesubsection{\metricStationary\ Agents.}
An agent is classified as stationary using an indicator function if the Euclidean distance between its start and end positions is less than a threshold $\delta_{\text{stat}} = 0.2$ meters:
\begin{equation*}
    \mathbb{I}_{\text{stat}}(i) = \begin{cases} 1 & \text{if } \| \mathbf{p}_{k_{\text{end}}^i}^i - \mathbf{p}_{k_{\text{start}}^i}^i \|_2 < \delta_{\text{stat}} \\ 0 & \text{otherwise} \end{cases}
\end{equation*}

\inlinesubsection{\metricFlow.}
To compute $\mflow$ and $\mflowEMD$, we explicitly define the local density calculation and directional partitioning.
For each agent $i$ at time $k$, local density $\rho_k^i$ (agents/m$^2$) is estimated using the Euclidean distance $r_k$ to its $K=4$th nearest neighbor: ${\rho_k^i = \frac{K}{\pi r_k^2}}$.

We partition all agent-timestep pairs $(i, k)$ into two sets based on the primary direction of movement:
\begin{itemize}
    \item $\mathcal{S}_x$: The set of pairs where movement is predominantly along the x-axis, $|v_{k,x}^i| > |v_{k,y}^i|$.
    \item $\mathcal{S}_y$: The set of pairs where movement is predominantly along the y-axis, $|v_{k,y}^i| > |v_{k,x}^i|$.
\end{itemize}
The instantaneous flow is $f_k^i = \rho_k^i \cdot \| \mathbf{v}_k^i \|_2$. The metric distributions $\mathcal{F}_x$ and $\mathcal{F}_y$ are constructed from the set of all $f_k^i$ belonging to $\mathcal{S}_x$ and $\mathcal{S}_y$, respectively.

\inlinesubsection{Nearest Neighbor Distribution.}
The nearest neighbor search includes a velocity threshold to exclude static agents. The metric is computed over all agent-timestep pairs $(i, k)$ where the agent $i$ is moving (speed $s_k^i > \epsilon=0.1$m/s). For each moving agent, we find its nearest moving neighbor, $j^*$, within a 10m radius:
$$
j^* = \argmin_{j \in \mathcal{A}_k, j \neq i, s_k^j > \epsilon} \|\mathbf{p}_k^i - \mathbf{p}_k^j\|_2
$$
The distance $\|\mathbf{p}_k^i - \mathbf{p}_k^{j^*}\|_2$ constitutes a sample in the distribution $D_{\text{nn}}$.

\subsection{Video Fidelity}

\inlinesubsection{\metricDisappearance.}
The central area of the frame described in Section \ref{metric:disappearance} is defined with a proportion $\alpha = 0.10$ of the frame dimensions. For a frame of width $W$ and height $H$ (in pixels), the edge margins are $m_x = \alpha W$ and $m_y = \alpha H$.
For each agent $i$, let $\mathbf{p}_{k_{\text{end}}^i}^i = (x_{k_{\text{end}}^i}^i, y_{k_{\text{end}}^i}^i)$ be its last detected position. An agent is classified as having disappeared mid-scene using an indicator function:
\begin{equation*}
\mathbb{I}_{\text{disp}}(i) = \begin{cases} 
1 & \text{if } m_x < x_{k_{\text{end}}^i}^i < W - m_x \text{ and} \\ 
  & \quad m_y < y_{k_{\text{end}}^i}^i < H - m_y \\ 
0 & \text{otherwise} 
\end{cases}
\end{equation*}

\section{T2V Prompt Suite}
\label{sec:prompt_suite}

\begin{figure}[b]
    \centering
    \includegraphics[width=0.75\linewidth]{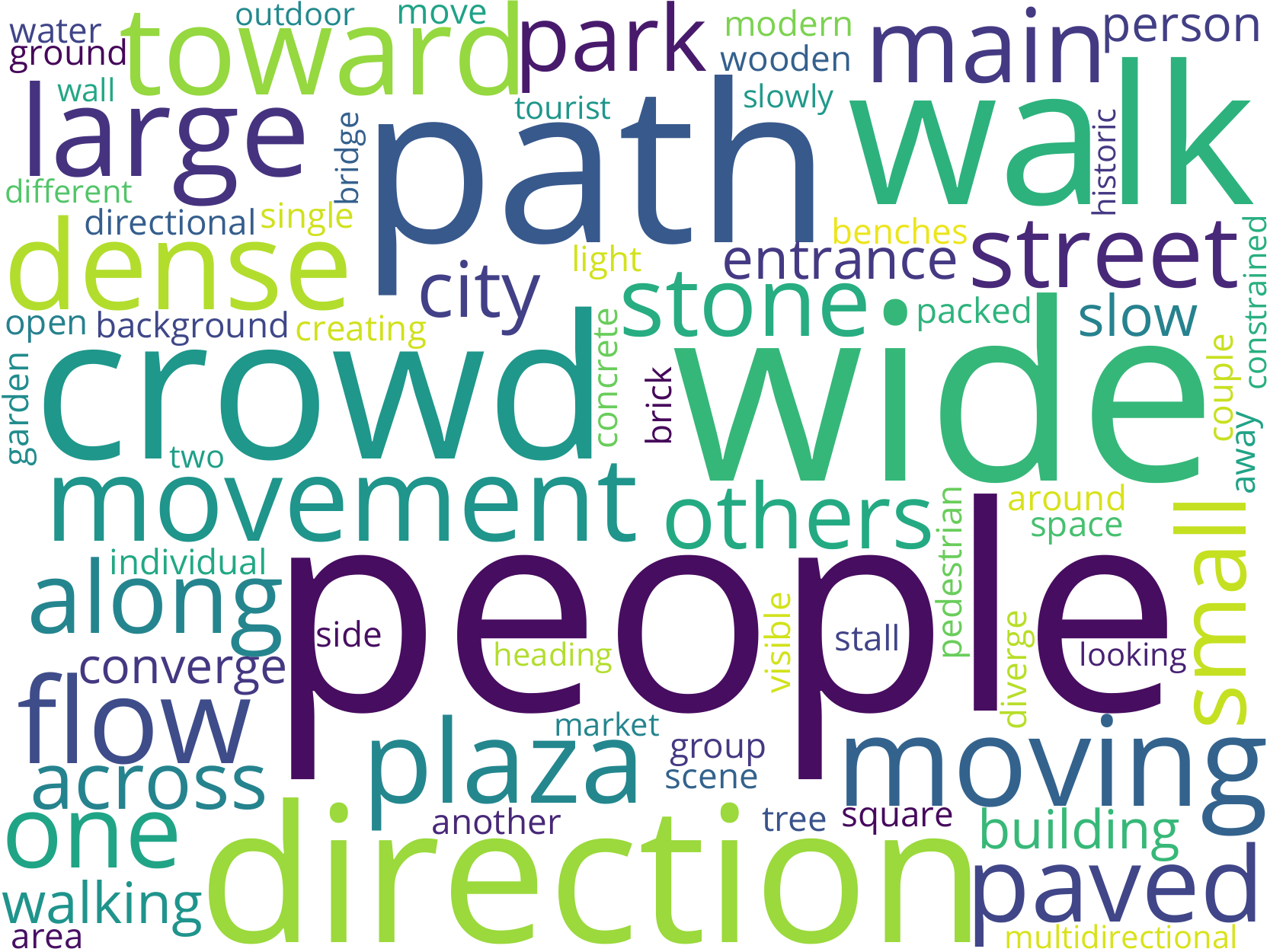}
    \caption{Word cloud visualization of prompt keywords in the T2V prompt suite.}
    \label{fig:prompts_word_cloud_sparse}
\end{figure}

\begin{figure*}[tb]
    \centering
    \includegraphics[width=0.95\linewidth]{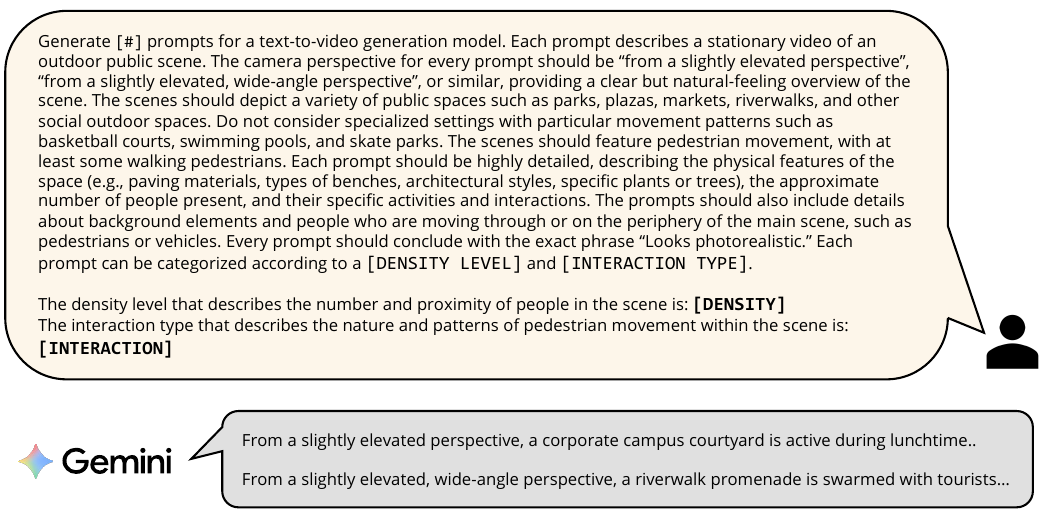}
    \caption{Script of the LLM instructions for generating text-to-video prompts. The instructions request a specific scene type, density, and interaction type, providing a standardized and compositional way to generate prompts. We used Gemini 2.5 Pro to generate the 180 prompts included in the supplementary material.}
    \label{fig:metaprompt}
\end{figure*}

The \emph{T2V Benchmark Method} section gave an overview of the method to systematically generate T2V prompts. Figure \ref{fig:prompts_word_cloud_sparse} visualizes the word cloud of the set of prompts showing the most common word is ``people,'' which distinctly contrasts with the prompt suite in Vbench \cite{huang_vbench_2024} where the most common word is ``person'' (singular).
In GRADEO \cite{mou2025gradeo} the most common human-oriented words are ``individual, person, boy, girl, man, woman.''
Figure \ref{fig:metaprompt} provides the full instruction script used to generate prompts by pasting in the desired density and interaction category. Other than the specific category strings, the instruction remains constant.

\section{Models and Inference Details}
\label{sec:inference_details}

\inlinesubsection{Model Selection.}
We selected five SOTA models that have both I2V and T2V variants: Wan2.1 (\emph{\WAN}) \cite{wan_wan_2025}, CogVideoX1.5 (\emph{\CVX}) \cite{yang_cogvideox_2025}, HunyuanVideo (\emph{\HYV}) \cite{kong_hunyuanvideo_2025}, LTX-Video (\emph{\LTX}) \cite{HaCohen2024LTXVideo}, and Open-Sora 2.0 (\emph{\OS}) \cite{peng2025opensora20}. The number of frames, video duration, video resolution, and frames per second (fps) vary by model due to the specific characteristics of the model architecture. Details of the model versions and generated video characteristics are provided in Table \ref{tab:video_models}. While some of these models support a variable number of generated frames, Open-Sora 2.0 and HunyuanVideo are capped at 129 frames as of the time of writing. 
Therefore, we choose to generate all clips at (approx.) 5 second durations, with the resolution as close as possible to the original resolution of the input image for I2V, or as close as possible to 720p for T2V.

\begin{table}[b]
    \centering
    \small
    \setlength{\tabcolsep}{.3mm}
    \caption{Video Generation Specifications.}
    \label{tab:video_models}
    \begin{tabular}{@{}lcccc@{}}
    \toprule
    \textbf{Model} & \textbf{FPS} & \textbf{Frames} & \textbf{Resolution} & \textbf{Duration (s)} \\
    \midrule
    \multicolumn{5}{l}{\textbf{Image-to-Video (I2V)}} \\
    CogVideoX1.5-5B-I2V & 16 & 81 & 640$\times$480* & 5.0 \\
    hunyuan-video-i2v-720p & 25 & 129 & 960$\times$540 & 5.12 \\
    ltxv-13b-0.9.7-dev & 30 & 153 & Same as input & 5.07 \\
    Open-Sora 2.0 768px & 24 & 129 & 880$\times$656 & 5.33 \\
    Wan2.1-I2V-14B-480P & 16 & 81 & 832$\times$480 & 5.0 \\
    \midrule
    \multicolumn{5}{l}{\textbf{Text-to-Video (T2V)}} \\
    CogVideoX1.5-5B & 16 & 81 & 1360$\times$768 & 5.0 \\
    hunyuan-video-t2v-720p & 25 & 129 & 1280$\times$720 & 5.12 \\
    ltxv-13b-0.9.7-dev & 30 & 153 & 1216$\times$704 & 5.07 \\
    Open-Sora 2.0 768px & 24 & 129 & 1024$\times$576 & 5.33 \\
    Wan2.1-T2V-14B & 16 & 81 & 1280$\times$720 & 5.0 \\
    \bottomrule
    \end{tabular}
    \caption*{\\ *CogVideoX I2V resolution varies by dataset:\\ETH (640$\times$480), UCY (720$\times$576)}
\end{table}

\inlinesubsection{\\Hyperparameters.} 
We keep the default or suggested hyperparameters for each of the models in order to match the best performance as reported by the authors. We use 50 inference steps for all models that take this as an input argument and otherwise leave the default. We specify \texttt{guidance\_scale=6.0} for CogVideoX, \texttt{guidance=7.5} and \texttt{guidance\_img=3.0} for Open-Sora 2.0, \texttt{embedded-cfg-scale=6.0} for HunyuanVideo, and \texttt{guidance\_scale=5.0} for Wan2.1, which were all chosen by referencing examples in the model \texttt{README} or default values in provided sample generation scripts. For HunyuanVideo, we manually adjusted \texttt{cfg-scale=1.2} up from 1.0 as the default value of 1.0 causes the inference script to disregard a negative prompt.

We generate random seeds for the video model in order to vary the generations using the same prompt. We store the seed in the filename for future reproducibility.

\inlinesubsection{\\Compute.}
In all generations, parallel inference was performed across four NVIDIA H200 GPUs, which resulted in generation times varying between 2 and 8 minutes per video depending on the model. Each GPU has 141GB memory, and we use a Linux machine with 16 CPUs and 128GB RAM. We use CUDA 12.4 and install all video generation models in conda environments according to the \texttt{README}.

\begin{table*}[t]
    \centering
    \footnotesize
    \setlength{\tabcolsep}{1mm}
    \begin{minipage}[t]{0.49\textwidth}
        \centering
        \caption{T2V Dataset Statistics: Number of Detections (N.D.), Number of Unique Agents (N.U.), and Average Number of Detections per Frame (D/F).}
        \label{tab:t2v_dataset_comprehensive_stats}
        \begin{tabular}{llrrrrrrr}
        \toprule
        Model & Count & Total & \multicolumn{3}{c}{Density} & \multicolumn{3}{c}{Interaction} \\
        \cmidrule(lr){4-6} \cmidrule(lr){7-9}
         & & & \denC & \denM & \denS & \intD & \intM & \intC \\
        \midrule
        \textbf{WAN} & N.D. & 3.88e6 & 3.19e6 & 5.75e5 & 1.08e5 & 1.12e6 & 1.58e6 & 1.18e6 \\
         & N.U. & 92431 & 79112 & 11367 & 1952 & 27148 & 39674 & 25609 \\
         & D/F & 56.83 & 136.69 & 25.51 & 4.86 & 49.65 & 67.92 & 52.52 \\
        \midrule
        \textbf{HYV} & N.D. & 3.76e6 & 2.58e6 & 1.02e6 & 1.65e5 & 1.22e6 & 1.36e6 & 1.19e6 \\
         & N.U. & 68594 & 48871 & 17168 & 2555 & 22143 & 26297 & 20154 \\
         & D/F & 35.36 & 72.21 & 27.58 & 4.89 & 35.17 & 37.99 & 32.95 \\
        \midrule
        \textbf{OS} & N.D. & 2.47e6 & 1.55e6 & 7.69e5 & 1.53e5 & 6.93e5 & 1.05e6 & 7.24e5 \\
         & N.U. & 36571 & 23605 & 11164 & 1802 & 10317 & 15608 & 10646 \\
         & D/F & 22.79 & 42.13 & 20.73 & 4.43 & 19.40 & 28.52 & 20.27 \\
        \midrule
        \textbf{LTX} & N.D. & 4.20e6 & 2.91e6 & 1.06e6 & 2.34e5 & 1.28e6 & 1.53e6 & 1.40e6 \\
         & N.U. & 54816 & 36992 & 14646 & 3178 & 16753 & 20497 & 17566 \\
         & D/F & 33.76 & 66.86 & 24.83 & 6.13 & 31.22 & 36.68 & 33.32 \\
        \midrule
        \textbf{CVX} & N.D. & 1.90e6 & 1.30e6 & 5.30e5 & 74789 & 5.92e5 & 7.73e5 & 5.37e5 \\
         & N.U. & 59584 & 44315 & 13563 & 1706 & 18197 & 25386 & 16001 \\
         & D/F & 28.51 & 55.86 & 22.97 & 3.66 & 27.38 & 34.18 & 23.90 \\
        \bottomrule
        \end{tabular}
    \end{minipage}
    \hfill
    \begin{minipage}[t]{0.49\textwidth}
        \centering
        \caption{I2V Dataset Statistics: Number of Detections (N.D.), Number of Unique Agents (N.U.), and Average Number of Detections per Frame (D/F). Note that the Ground Truth (GT) is shown as the top three rows.} %
        \label{tab:dataset_comprehensive_stats}
        \begin{tabular}{llrrrrr}
        \toprule
        Model & Count & ETH & UNIV & HOTEL & ZARA1 & ZARA2 \\
        \midrule
        \textbf{GT} & N.D. & 1861 & 101471 & 30505 & 8970 & 31624 \\
         & N.U. & 224 & 2488 & 1142 & 353 & 788 \\
         & D/F & 1.43 & 19.08 & 2.08 & 1.77 & 3.74 \\
        \midrule
        \textbf{WAN} & N.D. & 4957 & 21984 & 5945 & 4593 & 8558 \\
         & N.U. & 311 & 905 & 504 & 220 & 368 \\
         & D/F & 1.74 & 8.73 & 1.49 & 1.78 & 2.21 \\
        \midrule
        \textbf{HYV} & N.D. & 3738 & 59513 & 10099 & 10278 & 14987 \\
         & N.U. & 204 & 1209 & 435 & 315 & 499 \\
         & D/F & 1.60 & 11.84 & 1.51 & 1.99 & 2.25 \\
        \midrule
        \textbf{OS} & N.D. & 3329 & 28684 & 1950 & 4238 & 8760 \\
         & N.U. & 246 & 602 & 184 & 159 & 317 \\
         & D/F & 1.25 & 5.47 & 1.04 & 1.28 & 1.79 \\
        \midrule
        \textbf{LTX} & N.D. & 15305 & 73268 & 13769 & 16181 & 36580 \\
         & N.U. & 547 & 1460 & 629 & 345 & 651 \\
         & D/F & 2.12 & 12.86 & 1.60 & 2.76 & 3.87 \\
        \midrule
        \textbf{CVX} & N.D. & 2446 & 22066 & 2002 & 1970 & 2002 \\
         & N.U. & 236 & 944 & 227 & 159 & 193 \\
         & D/F & 1.51 & 6.79 & 1.18 & 1.39 & 1.45 \\
        \bottomrule
        \end{tabular}
    \end{minipage}
\end{table*}

\inlinesubsection{\\Synthetic Datasets.}
Tables \ref{tab:t2v_dataset_comprehensive_stats} and \ref{tab:dataset_comprehensive_stats} document the statistics of the number of detected agents across the T2V and I2V benchmark, respectively. The total number of detections (N.D.) counts the total number of bounding boxes across all frames and all video clips. The number of unique agents (N.U.) counts the total number of unique identifiers assigned by the MOT model, where each track ID corresponds to a unique person tracked across multiple frames. The number of detections per frame (D/F) counts the average number of bounding boxes detected per frame.

As discussed in the \emph{Method} section, the T2V prompt suite generated 5 repetitions for each of 20 prompts in each of the nine density/interaction categories (combinations of density \{\denC, \denM, \denS\} with interaction \{\intD, \intM, \intC\}). The videos are discarded if there is not sufficient agreement between the depth maps estimated by VGGT and Depth Pro. We require at least 100 pixels per frame for scale estimation, out of which at least 30\% must be determined inliers, where the residual depth error after scaling is less than a threshold of 10\% of the median metric depth.
The discard rates per model for the T2V benchmark were: WAN: 16/900 (1.78\%), HYV: 21/900 (2.33\%), CVX: 25/900 (2.78 \%), LTX: 45/900 (5.0\%), OS: 29/900 (3.22\%).

For I2V, as mentioned in the \emph{I2V Benchmark} section of the paper, we extract non-overlapping start frames at 5-second intervals. We generate a single video for each start frame for each model. The goal is to develop a synthetic video dataset that is the same length and with the \emph{same start distribution} of pedestrians as the ground truth dataset. For example, if we generated multiple videos for certain start frames, the resulting distributions would be biased towards that moment in time compared to the true reference. In the event that a model fails to produce a stationary and/or trackable video generation, we retry for that start image up to 5 times and retain the first video generation that contains any tracked agents.

\section{Additional Quantitative Results}
\label{sec:quantitative_results}

\subsection{Image-to-Video}

\begin{figure*}[!tp]
    \centering
    \vspace{3em}
    \begin{minipage}{\linewidth}
        \centering
        \begin{subfigure}[b]{0.195\linewidth}
            \includegraphics[width=\linewidth]{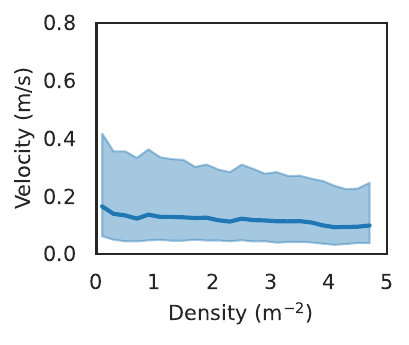}
            \caption*{CVX}
        \end{subfigure}
        \hfill
        \begin{subfigure}[b]{0.195\linewidth}
            \includegraphics[width=\linewidth]{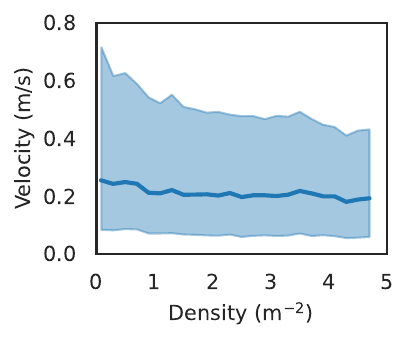}
            \caption*{HYV}
        \end{subfigure}
        \hfill
        \begin{subfigure}[b]{0.195\linewidth}
            \includegraphics[width=\linewidth]{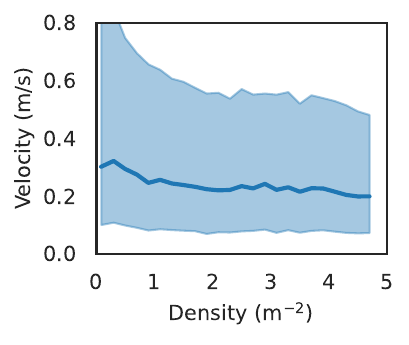}
            \caption*{LTX}
        \end{subfigure}
        \hfill
        \begin{subfigure}[b]{0.195\linewidth}
            \includegraphics[width=\linewidth]{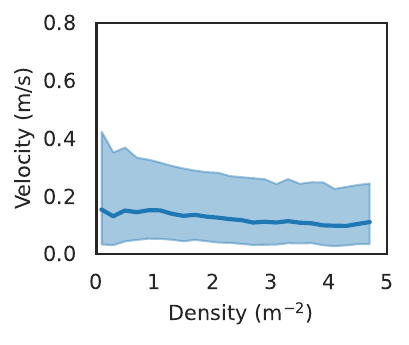}
            \caption*{OS}
        \end{subfigure}
        \hfill
        \begin{subfigure}[b]{0.195\linewidth}
            \includegraphics[width=\linewidth]{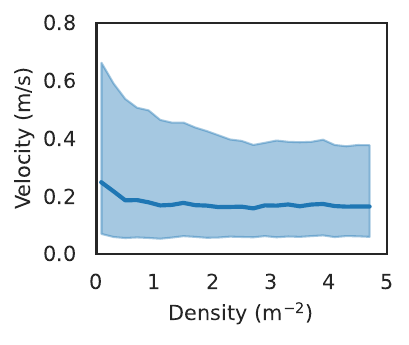}
            \caption*{WAN}
        \end{subfigure}
        \caption{Fundamental diagram plots for Crowded (\textbf{\denC}) pedestrian density in the T2V benchmark, showing the relationship between pedestrian flow and density simulated by different models. The center line and error bounds represent the median and Q1/Q3 quartiles. Plotting code adapted from Minartz et al. \cite{minartz_necs_2025}.}
        \label{fig:fd_crowded_tv}
    \end{minipage}
    
    \vspace{2em}
    
    \begin{minipage}{\linewidth}
        \centering
        \begin{subfigure}[b]{0.16\linewidth}
            \centering
            \includegraphics[width=\linewidth]{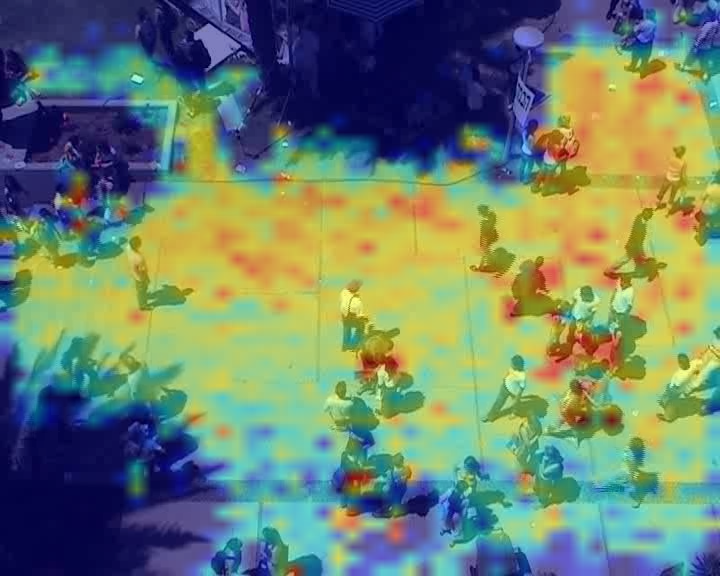}
            \caption*{GT}
        \end{subfigure}
        \begin{subfigure}[b]{0.16\linewidth}
            \centering
            \includegraphics[width=\linewidth]{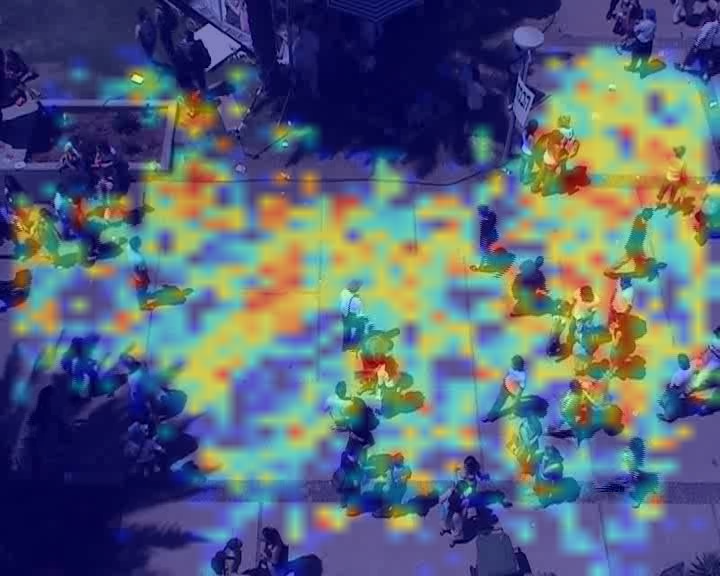}
            \caption*{CVX}
        \end{subfigure}
        \begin{subfigure}[b]{0.16\linewidth}
            \centering
            \includegraphics[width=\linewidth]{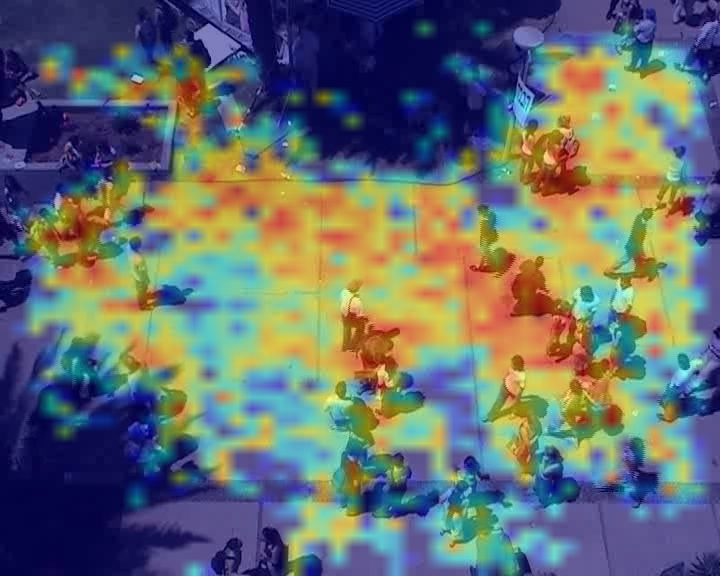}
            \caption*{HYV}
        \end{subfigure}
        \begin{subfigure}[b]{0.16\linewidth}
            \centering
            \includegraphics[width=\linewidth]{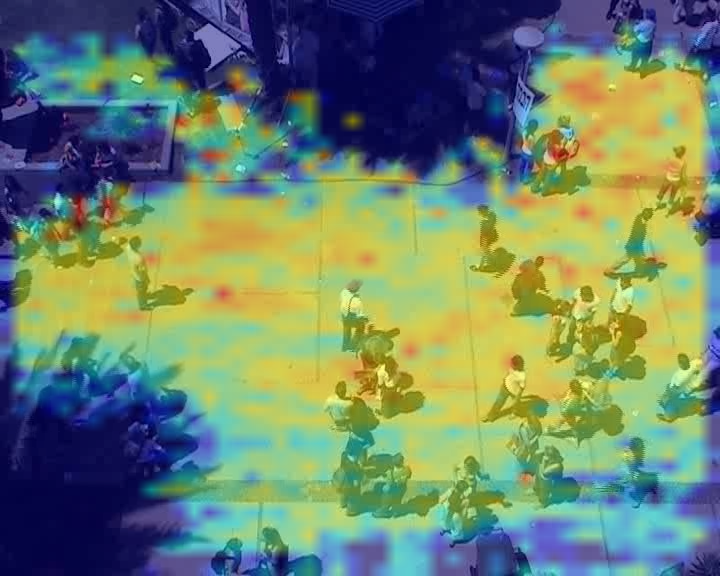}
            \caption*{LTX}
        \end{subfigure}
        \begin{subfigure}[b]{0.16\linewidth}
            \centering
            \includegraphics[width=\linewidth]{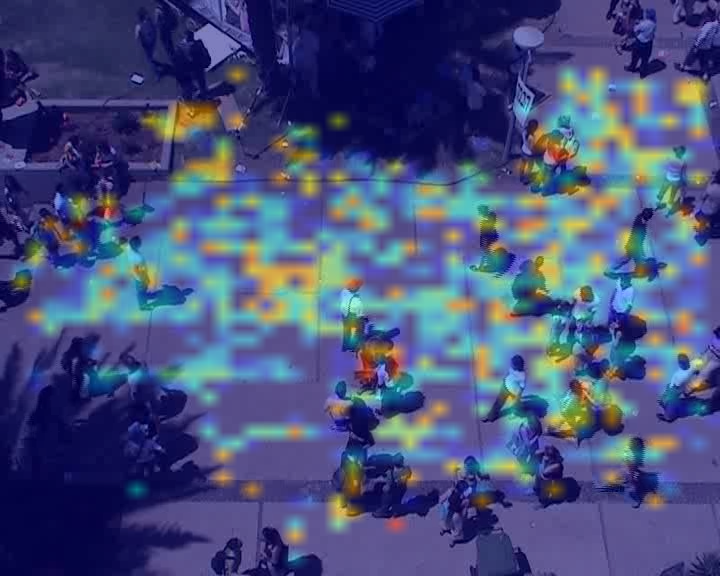}
            \caption*{OS}
        \end{subfigure}
        \begin{subfigure}[b]{0.16\linewidth}
            \centering
            \includegraphics[width=\linewidth]{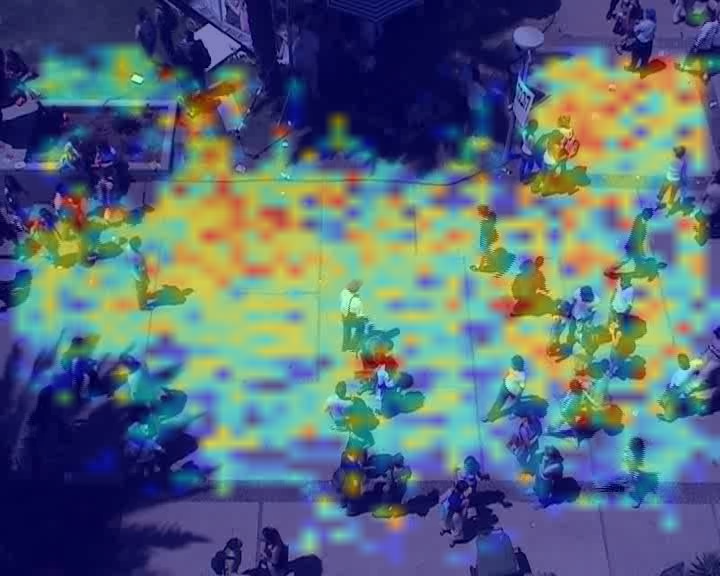}
            \caption*{WAN}
        \end{subfigure}
        \caption{2D histograms (heatmaps) of pedestrian positions for the UNIV scene from the I2V benchmark. Each subfigure shows the spatial distribution of pedestrian locations for ground truth and different models.}
        \label{fig:nn_heatmap_ucy}
    \end{minipage}
    
    \vspace{2em}
    
    \begin{minipage}{\linewidth}
        \centering
        \begin{subfigure}[b]{0.16\linewidth}
            \centering
            \includegraphics[width=\linewidth]{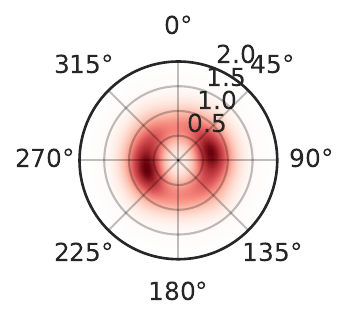}
            \caption*{GT}
        \end{subfigure}
        \begin{subfigure}[b]{0.16\linewidth}
            \centering
            \includegraphics[width=\linewidth]{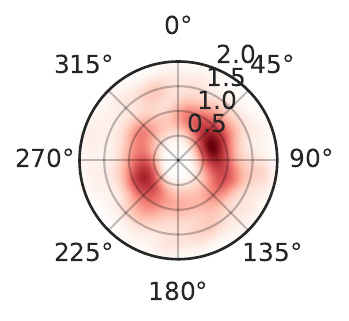}
            \caption*{CVX}
        \end{subfigure}
        \begin{subfigure}[b]{0.16\linewidth}
            \centering
            \includegraphics[width=\linewidth]{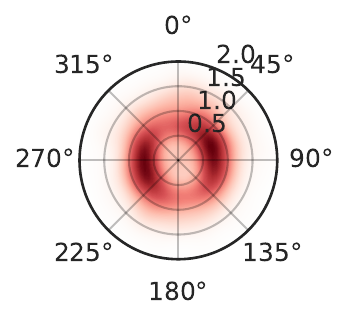}
            \caption*{HYV}
        \end{subfigure}
        \begin{subfigure}[b]{0.16\linewidth}
            \centering
            \includegraphics[width=\linewidth]{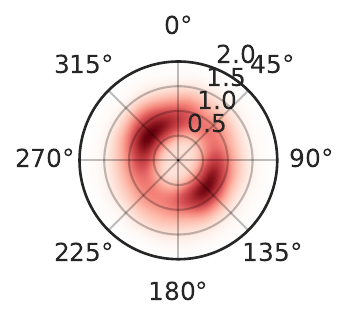}
            \caption*{LTX}
        \end{subfigure}
        \begin{subfigure}[b]{0.16\linewidth}
            \centering
            \includegraphics[width=\linewidth]{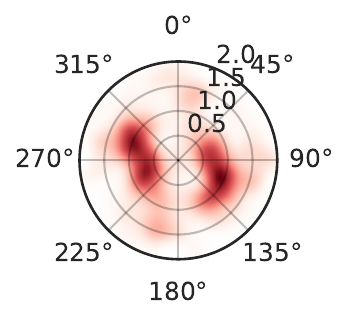}
            \caption*{OS}
        \end{subfigure}
        \begin{subfigure}[b]{0.16\linewidth}
            \centering
            \includegraphics[width=\linewidth]{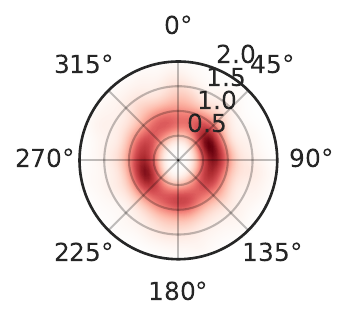}
            \caption*{WAN}
        \end{subfigure}
        \caption{Polar histograms of nearest neighbor (NN) relative positions for UNIV scene from the I2V benchmark. Each subfigure shows the angular distribution of the nearest neighbor with respect to the focused agent for different models and ground truth. Plotting code adapted form Minartz et al. \cite{minartz_necs_2025}.}
        \label{fig:nn_polar_ucy}
    \end{minipage}
\end{figure*}

\begin{table*}[htb!]
    \centering
    \label{tab:speed_results_simple}
    \setlength{\tabcolsep}{1.5mm}
    \small
    \vspace{3em}
    \caption{Mean agent speed (in m/s) $\pm$ standard deviation for non-stationary agents (displacement $>$ 0.2m).}
    \label{tab:rwkim}
    \begin{tabular}{llcccccc}
    \toprule
    Benchmark & Scene/Category & GT & \WAN & \HYV & \CVX & \LTX & \OS \\
    \midrule
    \multirow{5}{*}{\textbf{I2V}} & \textbf{ETH} & 1.29 $\pm$ 0.49 & 1.09 $\pm$ 0.61 & 1.93 $\pm$ 1.73 & 2.43 $\pm$ 1.55 & 1.78 $\pm$ 1.19 & 1.82 $\pm$ 1.13 \\
        & \textbf{HOTEL} & 1.38 $\pm$ 0.51 & 1.46 $\pm$ 0.78 & 1.05 $\pm$ 0.60 & 1.84 $\pm$ 1.16 & 1.82 $\pm$ 0.98 & 1.83 $\pm$ 0.86 \\
        & \textbf{UNIV} & 1.30 $\pm$ 0.65 & 1.03 $\pm$ 0.64 & 0.89 $\pm$ 0.57 & 1.01 $\pm$ 0.64 & 1.47 $\pm$ 0.72 & 0.92 $\pm$ 0.64 \\
        & \textbf{ZARA1} & 1.51 $\pm$ 0.49 & 1.27 $\pm$ 0.82 & 1.59 $\pm$ 0.95 & 2.04 $\pm$ 1.21 & 1.80 $\pm$ 0.95 & 1.95 $\pm$ 0.89 \\
        & \textbf{ZARA2} & 1.57 $\pm$ 0.58 & 1.04 $\pm$ 0.67 & 1.41 $\pm$ 0.53 & 1.87 $\pm$ 1.14 & 1.50 $\pm$ 0.65 & 1.55 $\pm$ 0.66 \\
    \midrule
    \multirow{7}{*}{\textbf{T2V}} & \textbf{\denS \intD} & --- & 0.89 $\pm$ 1.07 & 1.40 $\pm$ 1.32 & 0.64 $\pm$ 0.64 & 1.08 $\pm$ 1.16 & 0.73 $\pm$ 0.82 \\
    \cmidrule(lr){2-8}
    & {\denC} & --- & 0.53 $\pm$ 0.76 & 0.71 $\pm$ 0.97 & 0.42 $\pm$ 0.59 & 0.80 $\pm$ 0.99 & 0.40 $\pm$ 0.54 \\
    & {\denM} & --- & 0.61 $\pm$ 0.90 & 0.76 $\pm$ 1.04 & 0.47 $\pm$ 0.60 & 0.93 $\pm$ 1.07 & 0.46 $\pm$ 0.62 \\
    & {\denS} & --- & 0.71 $\pm$ 0.91 & 1.19 $\pm$ 1.40 & 0.61 $\pm$ 0.72 & 1.02 $\pm$ 1.18 & 0.62 $\pm$ 0.74 \\

    \cmidrule(lr){2-8}
    & {\intD} & --- & 0.76 $\pm$ 0.89 & 0.76 $\pm$ 0.94 & 0.48 $\pm$ 0.60 & 1.03 $\pm$ 1.14 & 0.53 $\pm$ 0.64 \\
        & {\intM} & --- & 0.61 $\pm$ 0.90 & 0.76 $\pm$ 1.04 & 0.47 $\pm$ 0.60 & 0.93 $\pm$ 1.07 & 0.46 $\pm$ 0.62 \\
        & {\intC} & --- & 0.53 $\pm$ 0.76 & 0.71 $\pm$ 0.97 & 0.42 $\pm$ 0.59 & 0.80 $\pm$ 0.99 & 0.40 $\pm$ 0.54 \\
    \bottomrule
    \end{tabular}
\end{table*}

\begin{table*}[!tp]
    \centering
    \small
    \vspace{4em}
    \caption{Complete I2V evaluation metrics.}
    \label{tab:i2v_results_full}
    \begin{tabular}{@{~~~~~}c@{~~~~~}c c@{~~~~~}c@{~~~~~}c@{~~~~~}c@{~~~~~}c@{~~~~~}c@{~~~~~}c@{~~~~~}c@{~~~~~}c@{~~~~~}c}
    \toprule
    \multirow{2}{*}{Dataset} & \multirow{2}{*}{Model} & \multicolumn{3}{c}{Trajectory Kinematics} & \multicolumn{5}{c}{Social Interaction} & \multicolumn{2}{c}{Fidelity} \\ \cmidrule(lr){3-5} \cmidrule(lr){6-10} \cmidrule(lr){11-12}
                           &               & $\mvelEMD$ $\downarrow$ & $\maccelEMD$ $\downarrow$ & $\mdistEMD$ $\downarrow$ & $\mcollEMD$ $\downarrow$ & $\mstatEMD$ $\downarrow$ & $\mpopEMD$ $\downarrow$ & $\mnnEMD$ $\downarrow$ & $\mflowEMD$ $\downarrow$ & $\mdisappearEMD$ $\downarrow$ & $\mmotconf$ $\uparrow$ \\ \midrule
    \cellcolor{gray!12} & \cellcolor{gray!12}WAN & \cellcolor{gray!12}\textbf{0.427} & \cellcolor{gray!12}\textbf{0.204} & \cellcolor{gray!12}\textbf{0.081} & \cellcolor{gray!12}0.001 & \cellcolor{gray!12}0.192 & \cellcolor{gray!12}0.360 & \cellcolor{gray!12}0.185 & \cellcolor{gray!12}1.331 & \cellcolor{gray!12}0.218 & \cellcolor{gray!12}0.476 \\
    \cellcolor{gray!12} & \cellcolor{gray!12}HYV & \cellcolor{gray!12}0.713 & \cellcolor{gray!12}1.947 & \cellcolor{gray!12}\underline{0.209} & \cellcolor{gray!12}\textbf{0.000} & \cellcolor{gray!12}\underline{0.157} & \cellcolor{gray!12}\underline{0.200} & \cellcolor{gray!12}0.333 & \cellcolor{gray!12}0.551 & \cellcolor{gray!12}\textbf{0.061} & \cellcolor{gray!12}\textbf{0.485} \\
    \cellcolor{gray!12} & \cellcolor{gray!12}OS & \cellcolor{gray!12}\underline{0.660} & \cellcolor{gray!12}1.785 & \cellcolor{gray!12}0.278 & \cellcolor{gray!12}\textbf{0.000} & \cellcolor{gray!12}\textbf{0.047} & \cellcolor{gray!12}0.211 & \cellcolor{gray!12}\underline{0.155} & \cellcolor{gray!12}\textbf{0.162} & \cellcolor{gray!12}\underline{0.092} & \cellcolor{gray!12}0.473 \\
    \cellcolor{gray!12} & \cellcolor{gray!12}LTX & \cellcolor{gray!12}0.815 & \cellcolor{gray!12}2.379 & \cellcolor{gray!12}0.841 & \cellcolor{gray!12}\underline{0.000} & \cellcolor{gray!12}0.318 & \cellcolor{gray!12}0.800 & \cellcolor{gray!12}\textbf{0.131} & \cellcolor{gray!12}3.104 & \cellcolor{gray!12}0.367 & \cellcolor{gray!12}0.478 \\
    \multirow{-5}{*}{\cellcolor{gray!12}ETH} & \cellcolor{gray!12}CVX & \cellcolor{gray!12}1.666 & \cellcolor{gray!12}\underline{0.489} & \cellcolor{gray!12}1.002 & \cellcolor{gray!12}0.006 & \cellcolor{gray!12}0.327 & \cellcolor{gray!12}\textbf{0.085} & \cellcolor{gray!12}0.329 & \cellcolor{gray!12}\underline{0.420} & \cellcolor{gray!12}0.143 & \cellcolor{gray!12}\underline{0.479} \\ \noalign{\vspace{1mm}}
    \multirow{5}{*}{UNIV} & WAN             & \textbf{0.383} & 1.027 & 0.554 & 0.081 & \underline{0.189} & 1.934 & 0.004 & 0.004 & 0.110 & 0.500 \\
                              & HYV             & 0.436 & \underline{0.525} & \underline{0.349} & \underline{0.063} & \textbf{0.021} & \underline{0.819} & \underline{0.001} & \underline{0.002} & 0.189 & \underline{0.505} \\
                              & OS              & 0.778 & 0.874 & 0.837 & 0.084 & 0.877 & 2.236 & 0.012 & 0.005 & 0.569 & 0.482 \\
                              & LTX             & \underline{0.402} & \textbf{0.240} & \textbf{0.245} & \textbf{0.029} & 0.223 & \textbf{0.715} & \textbf{0.001} & \textbf{0.001} & \textbf{0.074} & \textbf{0.510} \\
                              & CVX             & 0.434 & 0.968 & 0.603 & 0.111 & 0.301 & 2.297 & 0.004 & 0.005 & \underline{0.099} & 0.488 \\ \noalign{\vspace{1mm}}
    \cellcolor{gray!12} & \cellcolor{gray!12}WAN & \cellcolor{gray!12}\textbf{0.214} & \cellcolor{gray!12}0.696 & \cellcolor{gray!12}\underline{0.288} & \cellcolor{gray!12}\underline{0.013} & \cellcolor{gray!12}\textbf{0.016} & \cellcolor{gray!12}0.458 & \cellcolor{gray!12}0.067 & \cellcolor{gray!12}\textbf{0.052} & \cellcolor{gray!12}0.196 & \cellcolor{gray!12}\textbf{0.517} \\
    \cellcolor{gray!12} & \cellcolor{gray!12}HYV & \cellcolor{gray!12}\underline{0.402} & \cellcolor{gray!12}\underline{0.212} & \cellcolor{gray!12}0.369 & \cellcolor{gray!12}0.023 & \cellcolor{gray!12}0.175 & \cellcolor{gray!12}\underline{0.441} & \cellcolor{gray!12}\underline{0.048} & \cellcolor{gray!12}0.183 & \cellcolor{gray!12}\textbf{0.031} & \cellcolor{gray!12}\underline{0.499} \\
    \cellcolor{gray!12} & \cellcolor{gray!12}OS & \cellcolor{gray!12}0.459 & \cellcolor{gray!12}\textbf{0.108} & \cellcolor{gray!12}0.487 & \cellcolor{gray!12}0.023 & \cellcolor{gray!12}0.247 & \cellcolor{gray!12}0.815 & \cellcolor{gray!12}0.397 & \cellcolor{gray!12}0.209 & \cellcolor{gray!12}0.504 & \cellcolor{gray!12}0.498 \\
    \cellcolor{gray!12} & \cellcolor{gray!12}LTX & \cellcolor{gray!12}0.637 & \cellcolor{gray!12}0.406 & \cellcolor{gray!12}\textbf{0.048} & \cellcolor{gray!12}\textbf{0.011} & \cellcolor{gray!12}\underline{0.158} & \cellcolor{gray!12}\textbf{0.376} & \cellcolor{gray!12}\textbf{0.040} & \cellcolor{gray!12}\underline{0.097} & \cellcolor{gray!12}\underline{0.037} & \cellcolor{gray!12}0.494 \\
    \multirow{-5}{*}{\cellcolor{gray!12}HOTEL} & \cellcolor{gray!12}CVX & \cellcolor{gray!12}0.540 & \cellcolor{gray!12}0.632 & \cellcolor{gray!12}0.471 & \cellcolor{gray!12}0.023 & \cellcolor{gray!12}0.328 & \cellcolor{gray!12}0.703 & \cellcolor{gray!12}0.201 & \cellcolor{gray!12}0.209 & \cellcolor{gray!12}0.272 & \cellcolor{gray!12}0.499 \\ \noalign{\vspace{1mm}}
    \multirow{5}{*}{ZARA1} & WAN             & 0.603 & 0.947 & \underline{0.318} & \textbf{0.013} & 0.231 & \textbf{0.057} & 0.138 & 0.177 & 0.569 & 0.500 \\
                              & HYV             & \textbf{0.336} & \textbf{0.221} & \textbf{0.179} & \underline{0.034} & \textbf{0.134} & \underline{0.198} & 0.104 & 0.283 & 0.185 & \underline{0.503} \\
                              & OS              & 0.697 & \underline{0.264} & 0.320 & 0.056 & 0.386 & 0.425 & 0.165 & \underline{0.146} & 0.659 & 0.491 \\
                              & LTX             & \underline{0.569} & 0.557 & 0.571 & 0.086 & 0.239 & 0.866 & \underline{0.079} & 0.514 & \textbf{0.021} & \textbf{0.512} \\
                              & CVX             & 0.865 & 0.744 & 0.375 & 0.056 & \underline{0.209} & 0.330 & \textbf{0.076} & \textbf{0.104} & \underline{0.175} & 0.499 \\ \noalign{\vspace{1mm}}
    \cellcolor{gray!12} & \cellcolor{gray!12}WAN & \cellcolor{gray!12}0.658 & \cellcolor{gray!12}1.034 & \cellcolor{gray!12}0.610 & \cellcolor{gray!12}\textbf{0.037} & \cellcolor{gray!12}\underline{0.168} & \cellcolor{gray!12}0.698 & \cellcolor{gray!12}0.026 & \cellcolor{gray!12}0.356 & \cellcolor{gray!12}0.560 & \cellcolor{gray!12}0.494 \\
    \cellcolor{gray!12} & \cellcolor{gray!12}HYV & \cellcolor{gray!12}0.205 & \cellcolor{gray!12}\underline{0.291} & \cellcolor{gray!12}\underline{0.336} & \cellcolor{gray!12}\underline{0.044} & \cellcolor{gray!12}0.189 & \cellcolor{gray!12}\underline{0.679} & \cellcolor{gray!12}\underline{0.016} & \cellcolor{gray!12}\underline{0.120} & \cellcolor{gray!12}0.325 & \cellcolor{gray!12}\underline{0.506} \\
    \cellcolor{gray!12} & \cellcolor{gray!12}OS & \cellcolor{gray!12}\underline{0.150} & \cellcolor{gray!12}0.485 & \cellcolor{gray!12}0.386 & \cellcolor{gray!12}0.072 & \cellcolor{gray!12}\textbf{0.023} & \cellcolor{gray!12}0.891 & \cellcolor{gray!12}0.048 & \cellcolor{gray!12}0.377 & \cellcolor{gray!12}0.745 & \cellcolor{gray!12}0.486 \\
    \cellcolor{gray!12} & \cellcolor{gray!12}LTX & \cellcolor{gray!12}\textbf{0.128} & \cellcolor{gray!12}\textbf{0.154} & \cellcolor{gray!12}\textbf{0.250} & \cellcolor{gray!12}0.076 & \cellcolor{gray!12}0.375 & \cellcolor{gray!12}\textbf{0.085} & \cellcolor{gray!12}\textbf{0.004} & \cellcolor{gray!12}\textbf{0.007} & \cellcolor{gray!12}\textbf{0.150} & \cellcolor{gray!12}\textbf{0.520} \\
    \multirow{-5}{*}{\cellcolor{gray!12}ZARA2} & \cellcolor{gray!12}CVX & \cellcolor{gray!12}0.534 & \cellcolor{gray!12}0.695 & \cellcolor{gray!12}0.655 & \cellcolor{gray!12}0.075 & \cellcolor{gray!12}0.233 & \cellcolor{gray!12}1.046 & \cellcolor{gray!12}0.076 & \cellcolor{gray!12}0.400 & \cellcolor{gray!12}\underline{0.156} & \cellcolor{gray!12}0.492 \\
    \bottomrule
    \end{tabular}%
\end{table*}

\begin{table*}[p]
    \centering
    \small
    \setlength{\tabcolsep}{.8mm}
    \caption{T2V evaluation metrics across trajectory kinematics, social interaction, and video fidelity categories. The highest values are indicated in bold in order to emphasize trends between density and interaction categories. The real-world reference (Ref.) includes the ($\mu \pm \sigma$) of each metric computed across 10 public pedestrian benchmark datasets, as described in Section \ref{sec:eval}.} 
    \label{tab:t2v_evaluation}
    \begin{tabular}{cccccccccccccc}
    \toprule
    \multirow{3}{*}{Model} & \multirow{3}{*}{Category} & \multicolumn{3}{c}{Trajectory Kinematics} & \multicolumn{5}{c}{Social Interaction} & \multicolumn{3}{c}{Video Fidelity} \\ 
    \cmidrule(lr){3-5} \cmidrule(lr){6-10} \cmidrule(lr){11-13}
    & & $\mvel$ & $\maccel$ & $\mdist$ & $\mcollrate$ & $\mstat$ & $\mpop$ & $\mflow$ & $\mnnmode$ & $\mdisappear$ & $\mmotconf$ $\uparrow$ & $\mgeoconf$ $\uparrow$ \\
    & & (m/s) & (m/s$^2$) & (m) & (\%) & (\%) & (Count) & (1/m/s) & (m) & (\%) &  &  \\ \midrule
    & \textit{Ref.}  & \textit{.91 $\!\pm\!$ .84} & \textit{.65 $\!\pm\!$ .98} & \textit{3.62 $\!\pm\!$ 3.56} & \textit{1.19 $\!\pm\!$ 11.28} & \textit{.19 $\!\pm\!$ .39} & \textit{13.77 $\!\pm\!$ 19.07} & \textit{.54 $\!\pm\!$ .45} & \textit{1.18 $\!\pm\!$ 1.51} & - & - & - \\
    \cmidrule(lr){1-13}
    \multirow{7}{*}{\WAN} & \cellcolor{gray!12}\denC & \cellcolor{gray!12}\textbf{0.564} & \cellcolor{gray!12}\textbf{0.858} & \cellcolor{gray!12}\textbf{2.304} & \cellcolor{gray!12}\textbf{6.077} & \cellcolor{gray!12}0.199 & \cellcolor{gray!12}\textbf{136.693} & \cellcolor{gray!12}\textbf{1.541} & \cellcolor{gray!12}0.637 & \cellcolor{gray!12}\textbf{30.496} & \cellcolor{gray!12}0.531 & \cellcolor{gray!12}2.698 \\
     & \cellcolor{gray!12}\denM & \cellcolor{gray!12}0.556 & \cellcolor{gray!12}0.694 & \cellcolor{gray!12}1.327 & \cellcolor{gray!12}1.950 & \cellcolor{gray!12}0.330 & \cellcolor{gray!12}25.514 & \cellcolor{gray!12}0.230 & \cellcolor{gray!12}1.013 & \cellcolor{gray!12}25.732 & \cellcolor{gray!12}0.619 & \cellcolor{gray!12}4.687 \\
     & \cellcolor{gray!12}\denS & \cellcolor{gray!12}0.520 & \cellcolor{gray!12}0.554 & \cellcolor{gray!12}1.241 & \cellcolor{gray!12}1.272 & \cellcolor{gray!12}\textbf{0.338} & \cellcolor{gray!12}4.860 & \cellcolor{gray!12}0.029 & \cellcolor{gray!12}\textbf{1.806} & \cellcolor{gray!12}12.398 & \cellcolor{gray!12}\textbf{0.676} & \cellcolor{gray!12}\textbf{6.181} \\ \noalign{\vspace{1mm}}
     & \intC & 0.452 & 0.689 & 1.576 & \textbf{8.047} & \textbf{0.261} & 52.518 & 1.053 & 0.600 & 21.863 & \textbf{0.552} & 3.009 \\
     & \intD & \textbf{0.714} & \textbf{1.035} & \textbf{2.746} & 6.267 & 0.149 & 49.653 & \textbf{2.861} & 0.661 & 29.608 & 0.545 & 2.650 \\
     & \intM & 0.529 & 0.784 & 2.140 & 2.639 & 0.237 & \textbf{67.923} & 0.408 & \textbf{0.868} & \textbf{33.995} & 0.540 & \textbf{3.271} \\
    \cmidrule(lr){1-13}
    \multirow{7}{*}{\HYV} & \cellcolor{gray!12}\denC & \cellcolor{gray!12}0.553 & \cellcolor{gray!12}0.823 & \cellcolor{gray!12}1.517 & \cellcolor{gray!12}\textbf{11.926} & \cellcolor{gray!12}0.253 & \cellcolor{gray!12}\textbf{72.206} & \cellcolor{gray!12}\textbf{2.000} & \cellcolor{gray!12}0.582 & \cellcolor{gray!12}27.910 & \cellcolor{gray!12}0.526 & \cellcolor{gray!12}2.234 \\
     & \cellcolor{gray!12}\denM & \cellcolor{gray!12}0.897 & \cellcolor{gray!12}\textbf{1.131} & \cellcolor{gray!12}\textbf{1.712} & \cellcolor{gray!12}5.836 & \cellcolor{gray!12}0.256 & \cellcolor{gray!12}27.582 & \cellcolor{gray!12}1.015 & \cellcolor{gray!12}0.786 & \cellcolor{gray!12}\textbf{33.434} & \cellcolor{gray!12}0.584 & \cellcolor{gray!12}1.807 \\
     & \cellcolor{gray!12}\denS & \cellcolor{gray!12}\textbf{0.997} & \cellcolor{gray!12}1.055 & \cellcolor{gray!12}1.365 & \cellcolor{gray!12}2.749 & \cellcolor{gray!12}\textbf{0.289} & \cellcolor{gray!12}4.889 & \cellcolor{gray!12}1.912 & \cellcolor{gray!12}\textbf{1.597} & \cellcolor{gray!12}30.685 & \cellcolor{gray!12}\textbf{0.618} & \cellcolor{gray!12}\textbf{2.403} \\ \noalign{\vspace{1mm}}
     & \intC & 0.606 & 0.891 & 1.453 & \textbf{12.446} & 0.277 & 32.947 & 1.687 & 0.650 & 26.134 & \textbf{0.556} & 2.003 \\
     & \intD & \textbf{0.709} & \textbf{0.962} & \textbf{1.873} & 11.763 & 0.187 & 35.169 & \textbf{2.670} & 0.580 & 26.135 & 0.545 & 1.650 \\
     & \intM & 0.649 & 0.876 & 1.378 & 5.945 & \textbf{0.295} & \textbf{37.985} & 0.884 & \textbf{0.862} & \textbf{34.069} & 0.534 & \textbf{2.641} \\
    \cmidrule(lr){1-13}
    \multirow{7}{*}{\OS} & \cellcolor{gray!12}\denC & \cellcolor{gray!12}0.310 & \cellcolor{gray!12}0.466 & \cellcolor{gray!12}1.512 & \cellcolor{gray!12}\textbf{3.278} & \cellcolor{gray!12}0.269 & \cellcolor{gray!12}\textbf{42.129} & \cellcolor{gray!12}\textbf{0.245} & \cellcolor{gray!12}0.941 & \cellcolor{gray!12}\textbf{33.112} & \cellcolor{gray!12}0.535 & \cellcolor{gray!12}3.091 \\
     & \cellcolor{gray!12}\denM & \cellcolor{gray!12}\textbf{0.522} & \cellcolor{gray!12}\textbf{0.589} & \cellcolor{gray!12}\textbf{1.833} & \cellcolor{gray!12}1.909 & \cellcolor{gray!12}0.288 & \cellcolor{gray!12}20.731 & \cellcolor{gray!12}0.183 & \cellcolor{gray!12}1.156 & \cellcolor{gray!12}32.659 & \cellcolor{gray!12}0.585 & \cellcolor{gray!12}3.700 \\
     & \cellcolor{gray!12}\denS & \cellcolor{gray!12}0.477 & \cellcolor{gray!12}0.431 & \cellcolor{gray!12}1.738 & \cellcolor{gray!12}0.535 & \cellcolor{gray!12}\textbf{0.310} & \cellcolor{gray!12}4.428 & \cellcolor{gray!12}0.051 & \cellcolor{gray!12}\textbf{1.688} & \cellcolor{gray!12}12.875 & \cellcolor{gray!12}\textbf{0.657} & \cellcolor{gray!12}\textbf{4.591} \\ \noalign{\vspace{1mm}}
     & \intC & 0.310 & 0.432 & 1.302 & \textbf{4.124} & \textbf{0.329} & 20.268 & \textbf{0.283} & 0.883 & 28.161 & 0.559 & \textbf{4.008} \\
     & \intD & \textbf{0.476} & \textbf{0.590} & \textbf{2.111} & 2.324 & 0.193 & 19.401 & 0.239 & 0.984 & 30.445 & \textbf{0.562} & 2.378 \\
     & \intM & 0.371 & 0.491 & 1.515 & 1.927 & 0.297 & \textbf{28.519} & 0.143 & \textbf{1.229} & \textbf{35.277} & 0.550 & 3.546 \\
    \cmidrule(lr){1-13}
    \multirow{7}{*}{\LTX} & \cellcolor{gray!12}\denC & \cellcolor{gray!12}0.760 & \cellcolor{gray!12}\textbf{1.207} & \cellcolor{gray!12}\textbf{2.207} & \cellcolor{gray!12}\textbf{9.827} & \cellcolor{gray!12}0.212 & \cellcolor{gray!12}\textbf{66.856} & \cellcolor{gray!12}\textbf{1.489} & \cellcolor{gray!12}0.606 & \cellcolor{gray!12}27.200 & \cellcolor{gray!12}0.555 & \cellcolor{gray!12}1.411 \\
     & \cellcolor{gray!12}\denM & \cellcolor{gray!12}\textbf{0.904} & \cellcolor{gray!12}1.188 & \cellcolor{gray!12}1.706 & \cellcolor{gray!12}2.955 & \cellcolor{gray!12}0.287 & \cellcolor{gray!12}24.829 & \cellcolor{gray!12}0.448 & \cellcolor{gray!12}0.961 & \cellcolor{gray!12}41.704 & \cellcolor{gray!12}\textbf{0.574} & \cellcolor{gray!12}\textbf{1.563} \\
     & \cellcolor{gray!12}\denS & \cellcolor{gray!12}0.820 & \cellcolor{gray!12}1.039 & \cellcolor{gray!12}1.402 & \cellcolor{gray!12}1.364 & \cellcolor{gray!12}\textbf{0.317} & \cellcolor{gray!12}6.130 & \cellcolor{gray!12}0.164 & \cellcolor{gray!12}\textbf{1.458} & \cellcolor{gray!12}\textbf{42.259} & \cellcolor{gray!12}0.569 & \cellcolor{gray!12}1.403 \\ \noalign{\vspace{1mm}}
     & \intC & 0.648 & 1.004 & 1.597 & 9.011 & \textbf{0.303} & 33.325 & 1.006 & 0.718 & 30.121 & 0.563 & 1.433 \\
     & \intD & \textbf{0.966} & \textbf{1.370} & \textbf{2.518} & \textbf{9.094} & 0.162 & 31.218 & \textbf{1.819} & 0.652 & 30.663 & \textbf{0.569} & 1.258 \\
     & \intM & 0.799 & 1.208 & 1.992 & 5.114 & 0.245 & \textbf{36.675} & 0.717 & \textbf{0.837} & \textbf{35.020} & 0.552 & \textbf{1.625} \\
    \cmidrule(lr){1-13}
    \multirow{7}{*}{\CVX} & \cellcolor{gray!12}\denC & \cellcolor{gray!12}0.370 & \cellcolor{gray!12}0.629 & \cellcolor{gray!12}\textbf{1.222} & \cellcolor{gray!12}\textbf{6.760} & \cellcolor{gray!12}0.301 & \cellcolor{gray!12}\textbf{55.856} & \cellcolor{gray!12}\textbf{0.605} & \cellcolor{gray!12}0.702 & \cellcolor{gray!12}\textbf{34.178} & \cellcolor{gray!12}0.494 & \cellcolor{gray!12}\textbf{3.020} \\
     & \cellcolor{gray!12}\denM & \cellcolor{gray!12}0.468 & \cellcolor{gray!12}\textbf{0.698} & \cellcolor{gray!12}1.163 & \cellcolor{gray!12}3.803 & \cellcolor{gray!12}0.319 & \cellcolor{gray!12}22.971 & \cellcolor{gray!12}0.314 & \cellcolor{gray!12}0.926 & \cellcolor{gray!12}34.152 & \cellcolor{gray!12}0.553 & \cellcolor{gray!12}2.244 \\
     & \cellcolor{gray!12}\denS & \cellcolor{gray!12}\textbf{0.494} & \cellcolor{gray!12}0.644 & \cellcolor{gray!12}1.036 & \cellcolor{gray!12}2.642 & \cellcolor{gray!12}\textbf{0.322} & \cellcolor{gray!12}3.664 & \cellcolor{gray!12}0.088 & \cellcolor{gray!12}\textbf{1.546} & \cellcolor{gray!12}20.340 & \cellcolor{gray!12}\textbf{0.598} & \cellcolor{gray!12}2.053 \\ \noalign{\vspace{1mm}}
     & \intC & 0.333 & 0.546 & 0.912 & \textbf{7.299} & \textbf{0.382} & 23.899 & 0.470 & 0.816 & 30.486 & \textbf{0.519} & \textbf{3.033} \\
     & \intD & \textbf{0.441} & \textbf{0.719} & \textbf{1.511} & 6.227 & 0.234 & 27.381 & \textbf{0.617} & 0.746 & 30.829 & 0.513 & 2.541 \\
     & \intM & 0.402 & 0.655 & 1.166 & 4.367 & 0.309 & \textbf{34.175} & 0.425 & \textbf{0.835} & \textbf{37.777} & 0.503 & 2.876 \\
    \bottomrule
    \end{tabular}%
\end{table*}

Fig. \ref{fig:nn_heatmap_ucy} plots a heatmap of the agent positions on top of a background image of the UNIV scene. The plots show that all of the models roughly capture the shape of the ground truth (GT) spatial distribution. However, the relative densities vary. \LTX\ appears to capture the true distribution best. \CVX\ and \OS\ show sparser overall distributions due to the lower trackability of the agents, resulting in fewer detected pedestrians over the same number of generated video clips; we note that the \metricMOTConf\ metric ($\mmotconf$) captures this in Table \ref{tab:i2v_results_full} as the lowest two scores in the UNIV scene.

Figure \ref{fig:nn_polar_ucy} shows the polar histograms for the same scene, which illustrate the relative location of the nearest neighbor to each agent. Prior research has noted that this position typically follows a bimodal distribution with peaks at distances around 0.5-0.75 meters \cite{minartz_necs_2025}. 
The GT distribution indeed follows this pattern. 
Intuitively, this results from the fact that people walk side-by-side with some personal space in between themselves and the nearest other person. To some degree it reflects collision avoidance behavior, as two colliding walkers would lead to a nearest neighbor distance near zero, which would result in a dense cluster near the origin. Models \HYV\ and \WAN\ capture the GT NN distribution well, including the distance and angle of the two modes. \LTX\ captures the distance of the NN modes well but displays a different relative orientation angle. \CVX\ roughly captures the GT pattern but less clearly due to the sparser nature of detections resulting from lower agent trackability. \OS\ displays the largest visual dissimilarity against the GT distribution, which is accurately reflected in Table \ref{tab:i2v_results_full} by the worst score in the UNIV scene for the $\mnnEMD$ metric. 

\subsection{Text-To-Video}

\inlinesubsection{Details on Real-World Reference Datasets.}
To establish reference ranges for T2V evaluation, we process each of the ten public pedestrian benchmarks using dataset-specific loaders from the OpenTraj toolkit \cite{amirian2020opentraj}. For datasets with known camera parameters (ETH, UCY, Town Center, PETS-2009, WildTrack), we apply pre-computed homography matrices or calibration files to project pixel-space detections into metric bird's-eye view coordinates. For datasets lacking explicit camera models (Edinburgh, Grand Central, KITTI, HERMES), we use the native world coordinates provided by the original annotations.
We apply the same preprocessing pipeline used for tracking data produced from the video generation models, including 5-second temporal windowing, matching the duration of the synthetic video. We compute the full suite of trajectory kinematics, social interaction, and video fidelity metrics (Section~\ref{sec:eval}) on each processed dataset, aggregating results across all scenes to establish the reference distribution ranges reported in our evaluation.\\

\inlinesubsection{Details on Flow.}
As discussed in the paper, an inverse relationship is expected where increasing crowd density results in decreasing average walking speeds \cite{Seyfried_2005}. The Fruin level of service (LOS) \cite{fruin1971pedestrian} provides an intuitive understanding of different crowd densities: 
\begin{itemize}
    \item LOS A, $>$13 ft$^2$/ped ($<$0.83 ped/m$^2$)
    \item LOS B, 10-13 ft$^2$/ped (.83-1.08 ped/m$^2$)
    \item LOS C, 6-10 ft$^2$/ped (1.08-1.79 ped/m$^2$)
    \item LOS D, 3-6 ft$^2$/ped (1.79 - 3.59 ped/m$^2$)
    \item LOS E, 2-3 ft$^2$/ped (3.59 - 5.38 ped/m$^2$)
    \item LOS F, $<$2 ft$^2$/ped ($>$ 5.38 ped/m$^2$)
\end{itemize}
LOS A corresponds to free standing and circulation without disturbing others. LOS C corresponds to restricted circulation but still within the range of comfort. LOS E corresponds to serious discomfort where physical contact with others is unavoidable.

Figure \ref{fig:fd_crowded_tv} shows the fundamental diagrams for the Croweded (\denC) density of the T2V benchmark. The maximum density on these plots of 5 people per sq. m results in shoulder-to-shoulder spacing with highly restricted movement. All five T2V models roughly capture the expected decreasing trend. However, the decrease in walking speed tends to plateau for all models above 2-3 ped/m$^2$, which does not reflect the expected behavior.\\

\inlinesubsection{Real-World Interpretation of Velocity.}

The $\mvel$ and $\mvelEMD$ metrics reported in the main paper include all pedestrians in the scene. Since each scene contains some percentage of stationary pedestrians (given by $\mstat$), this decreases the average walking speed. Here we analyze the walking speeds of only agents that have a non-zero overall displacement in order to give a more intuitive analysis of how realistic the pace is (Table \ref{tab:rwkim}).

A number of peer-reviewed studies report statistics on the distributions of human walking speeds in unobstructed environments, i.e., not restricted by the presence of other humans or obstacles \cite{BOHANNON2011182,syddall_2015,mohler_visual_2007}. They range from 0.8 m/s for elderly populations to as high as 1.6 m/s for healthy adult males, with an average walking speed reported around 1.3 m/s. For I2V, we can compute the ground truth walking speeds as a point of comparison. Table \ref{tab:rwkim} reports the results on the GT datasets from the ETH/UCY scenes. The walking speeds range from 1.29 m/s (ETH) to 1.57 m/s (ZARA2), which strongly agree with the results expected from the literature. The Sparse/Directional subset (\emph{\denS/\intD}) the T2V benchmark corresponds to the same type of scenery as ETH/UCY and serves as a good point of comparison. 

In the I2V benchmark, \WAN\ is the model that produces the closest match overall to GT walking speed distributions. The other models have variable performance, with some scenes very close to the GT distribution and others clearly too fast or too slow, although still within a range of physically plausible movement speeds.  Speeds as high as 2.43 m/s (\CVX, ETH scene) approach running speeds rather than walking, which appears to result from a scale mismatch where the model generates humans that are too large relative to the scene and therefore walk too quickly in the real-world coordinate system.

In the T2V benchmark, \HYV\ is the model that closest approximates the expected speed distribution, averaging 1.40 m/s in the \denS/\intD category. All of the other models produce \emph{walking speeds which are generally too slow} (especially \CVX), although the standard deviation is high enough that many pedestrians fall within the normal range. This result is especially interesting given the feasible walking speeds produced in the I2V benchmark.

\section{Additional Qualitative Results}
\label{sec:qualitative_results}
\subsection{T2V Scene Variety}
\begin{figure*}[p]
    \centering
    \begin{subfigure}{\linewidth}
        \includegraphics[width=\linewidth]{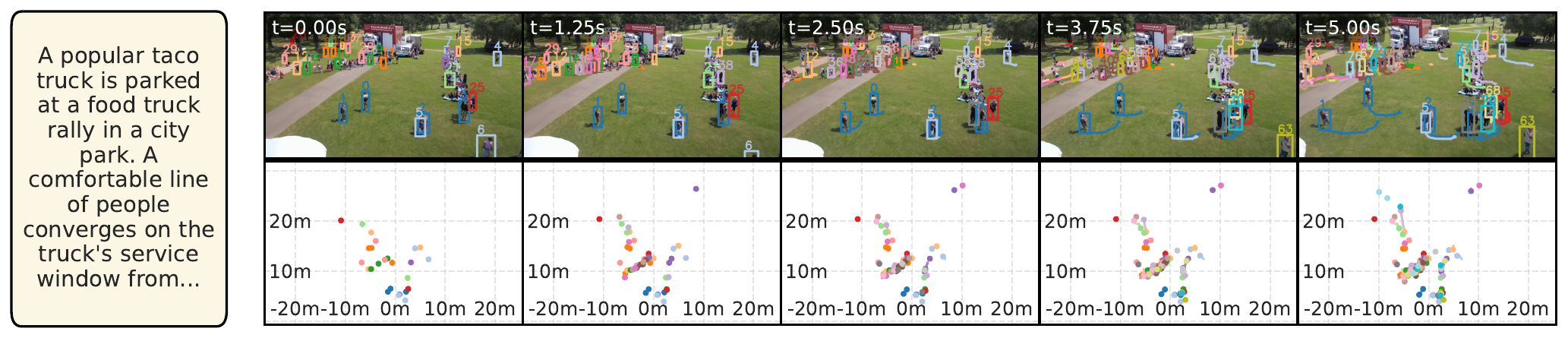}
        \caption{CogVideoX1.5 (\CVX), \intC}
    \end{subfigure}
    \\
    \begin{subfigure}{\linewidth}
        \includegraphics[width=\linewidth]{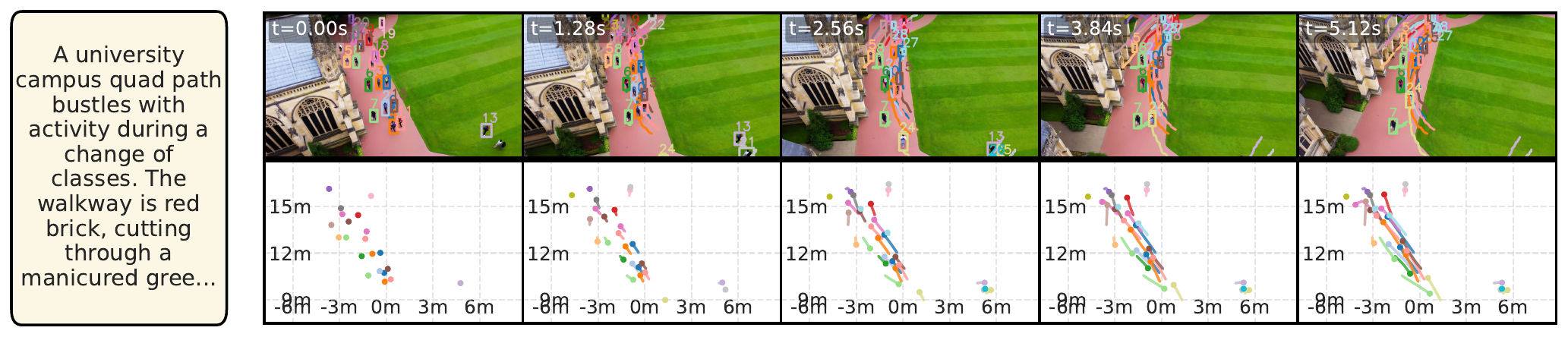}
        \caption{HunyuanVideo (\HYV), \intD}
    \end{subfigure}
    \\
    \begin{subfigure}{\linewidth}
        \includegraphics[width=\linewidth]{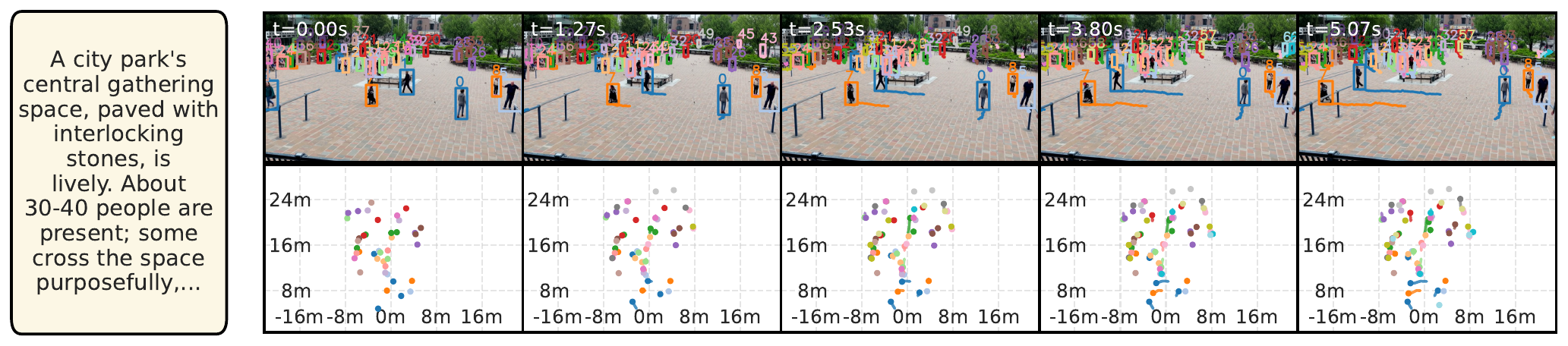}
        \caption{LTX-Video (\LTX), \intM}
    \end{subfigure}
    \\
    \begin{subfigure}{\linewidth}
        \includegraphics[width=\linewidth]{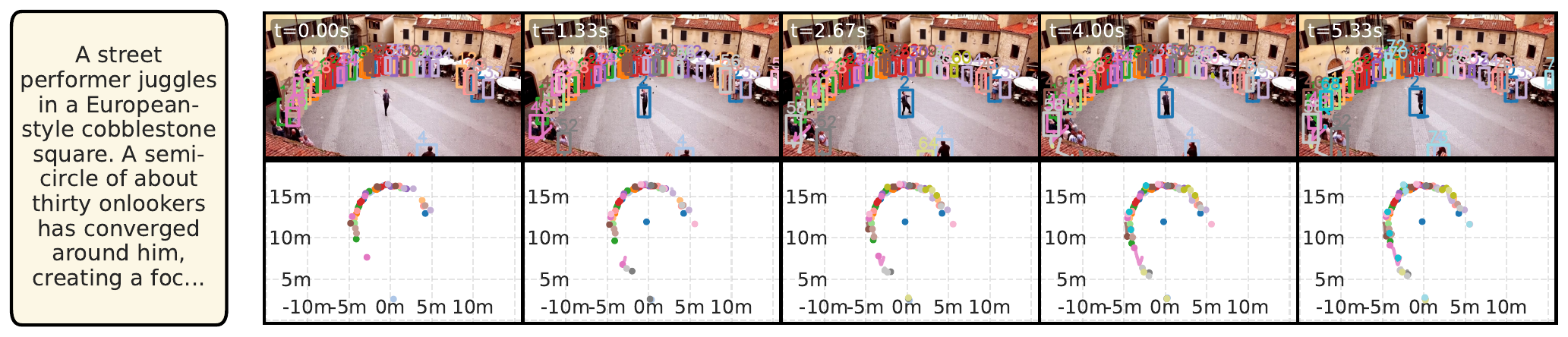}
        \caption{Open-Sora 2.0 (\OS), \intC}
    \end{subfigure}
    \\
    \begin{subfigure}{\linewidth}
        \includegraphics[width=\linewidth]{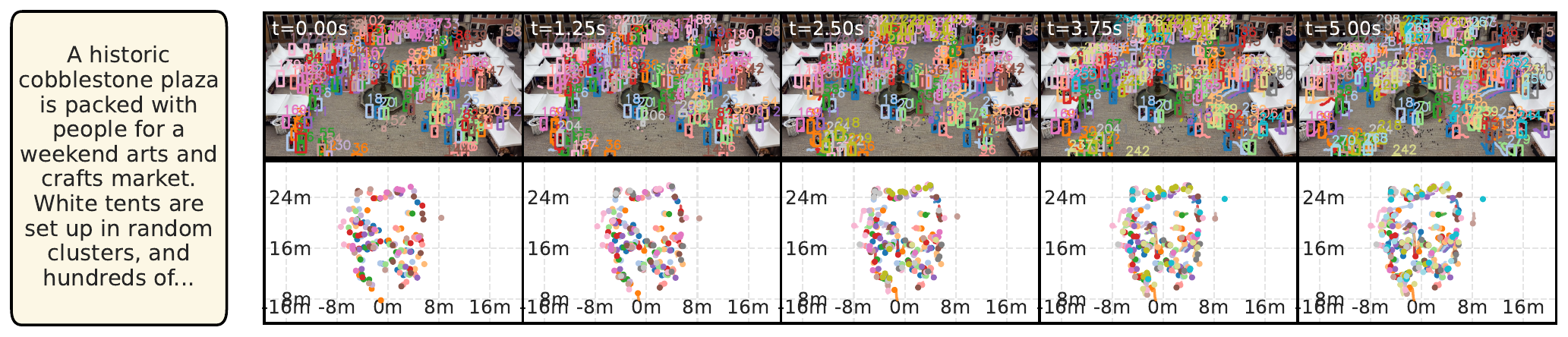}
        \caption{Wan2.1 (\WAN), \intM}
    \end{subfigure}
    \caption{Additional qualitative examples showing a variety of specified interaction types from the T2V prompt suite.}
    \label{fig:qualitative_examples}
\end{figure*}

Figure \ref{fig:qualitative_examples} shows additional examples of trajectory extraction and BEV coordinates from T2V generations by various models with all three interaction types. We note the high degree of success of the multi-object tracking and the realistic metric scales computed using the process described in the \emph{Method} section. Figure (b) demonstrates that even with large degrees of camera motion, the use of frame-wise camera extrinsics from VGGT allows a consistent world coordinate system to be established such that 1) the walking trajectories remain aligned on a \emph{straight line following the red path}, despite the pixel-coordinate paths taking on a curve due to the camera motion; and 2) the seated people in the bottom right corner retain stationary locations. Figure (d) demonstrates the significant scene and behavior variety that can be obtained through text prompts alone, especially in scenes that would be challenging or impossible to specify in conventional simulation software. Figure (e) demonstrates that crowded scenes with over 100 pedestrians remain successfully tracked, showing the power of this method to extract large numbers of trajectories in a single generation.

\subsection{Failure Modes}

\begin{figure*}[p]
    \centering
    \includegraphics[width=0.95\linewidth]{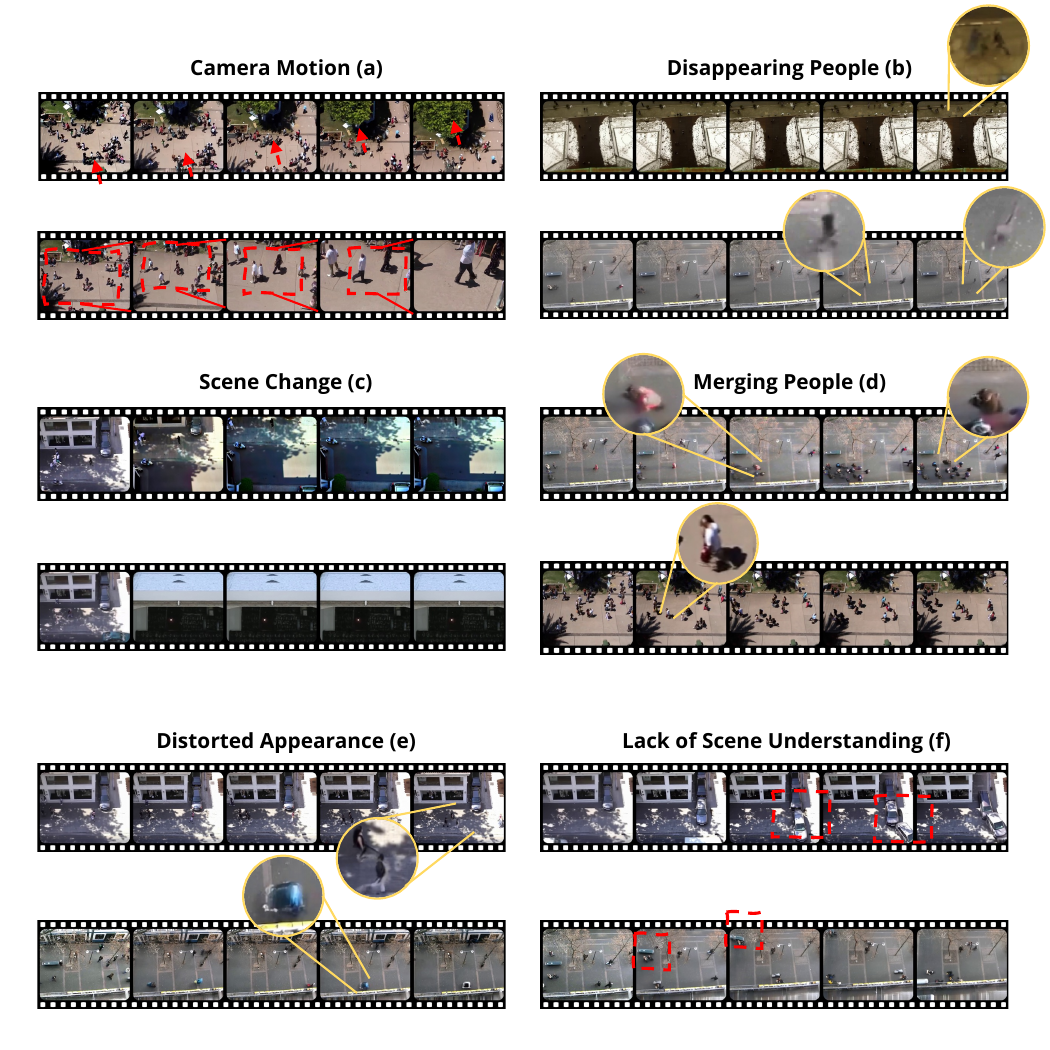}
    \caption{Common failure modes observed in I2V generations. (a) The camera perspective may pan (top) or zoom (bottom) despite the request for a stationary view in the positive and negative prompts. We filter out these videos as it prevents using the ETH/UCY homography matrices. (b) People spontaneously disappear from one frame to another or become ghostly. (c) The scene many abruptly change despite starting off as the scene given by the image condition. (d) People who begin as separate agents may merge into one another, which results in disappearing MOT track IDs. (e) People may have elongated or distorted appearance (top). Objects may appear that do not look like either people or vehicles (bottom). (f) A car at the curb which should remain parked moves forward as if driving on a road (top); a bench begins to move as if it is some type of vehicle (bottom). }
    \label{fig:i2v_failure_modes}
\end{figure*}

\begin{figure*}[p]
    \centering
    \includegraphics[width=0.95\linewidth]{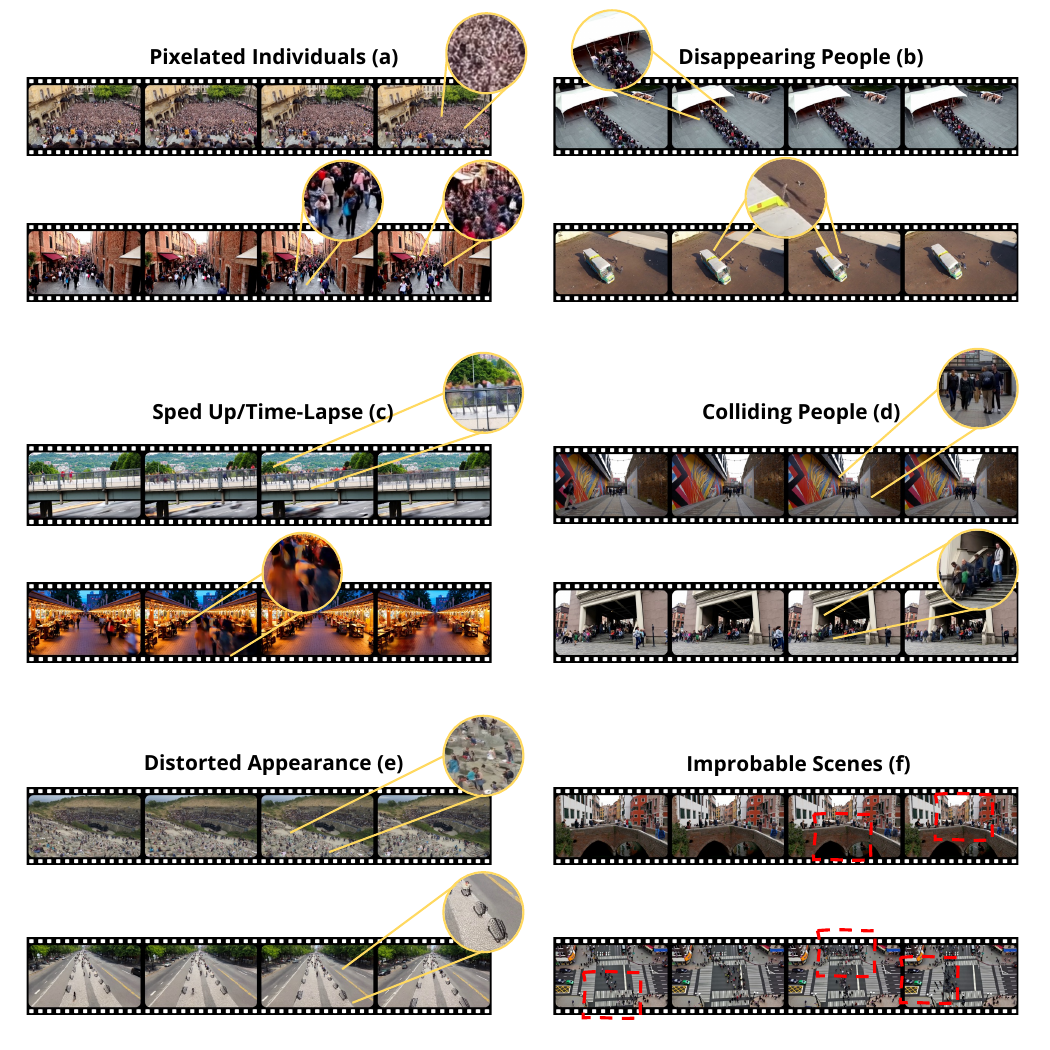}
    \caption{Common failure modes observed in T2V generations. (a) Individuals lose trackability in crowds when the depiction turns into a pixelated mass, which is more prominent for far-away people in the background than close-up people represented by more pixels. (b) Pedestrians in a dense queue disappear as they move through a bottleneck rather than re-emerging on the other side (top); an individual pedestrian in a sparse scene disappearing (bottom). (c) Undesired sped-up or time-lapse effects in generated videos cause high blurring in individual frames which prevents tracking. (d) While colliding pedestrians are more common in dense pedestrian flows (bottom), there are also examples where individuals walk directly into oncoming groups (top). (e) Scenes and/or people may have distorted appearances, which impacts the success of 3D reconstruction and tracking, respectively. (f) Scenes may be physically improbable, both in terms of 3D space (top, ill-defined perspective) or context (bottom, duplicate crosswalks).}
    \label{fig:t2v_failure_modes}
\end{figure*}

Figures \ref{fig:i2v_failure_modes} and \ref{fig:t2v_failure_modes} illustrate examples of failure modes for the image-to-video and text-to-video benchmarks, respectively.\\

\inlinesubsection{Common Failure Modes}
\begin{itemize}
    \item Disappearing Pedestrians (Figures \ref{fig:i2v_failure_modes}b and \ref{fig:t2v_failure_modes}b): One of the most prevalent issues is the spontaneous vanishing of pedestrians mid-trajectory .
    \item Merging/Colliding People  (Figures  \ref{fig:i2v_failure_modes}d and \ref{fig:t2v_failure_modes}d): Rather than exhibiting realistic collision avoidance behavior, pedestrians frequently merge together or occupy the same spatial location.
    \item Visual Distortions (Figures  \ref{fig:i2v_failure_modes}e and \ref{fig:t2v_failure_modes}e): Degradation in pedestrian appearance may render individuals unrecognizable or untrackable by the multi-object tracker. Distorted objects that are neither pedestrian nor vehicle sometimes appear.
\end{itemize}

\inlinesubsection{\\I2V-Specific Failure Modes}
\begin{itemize}
    \item Unwanted Camera Motion (Figure \ref{fig:i2v_failure_modes}a): Models may introduce camera movement despite static camera prompts. Since we use ETH/UCY pre-computed homography matrices, this represents a failure mode for the I2V benchmark, although the T2V benchmark is designed to expect camera motion.
    \item Scene Changes (Figure \ref{fig:i2v_failure_modes}c): Models may spontaneously change scene from the input image.
    \item Scene Understanding (Figure \ref{fig:i2v_failure_modes}f): Models may inappropriately animate static objects, such as moving parked cars in pedestrian-only zones. This suggests limitations in the latent representation of the input condition image.
\end{itemize}

\inlinesubsection{\\T2V-Specific Failure Modes}
\begin{itemize}
    \item Pixelated Masses (Figure \ref{fig:t2v_failure_modes}a): In crowded scenarios, models often fail to render distinct individuals, instead producing untrackable, fluid-like pixelated masses.
    \item Sped Up/Time-Lapse Effects (Figure \ref{fig:t2v_failure_modes}c): Models sometimes generate unwanted temporal acceleration, causing pedestrians to appear as motion blur streaks. This doesn't affect the velocity metrics ($\mvel, \mvelEMD$) as the blurred people are not detected by the MOT model.
    \item Improbable Scene Generation (Figure \ref{fig:t2v_failure_modes}f): T2V models may create impossible scenarios with inappropriate semantic context or 3D physicality.
\end{itemize}

\end{document}